\documentclass{article}

\usepackage[preprint]{rlc}

\usepackage[utf8]{inputenc}
\usepackage[T1]{fontenc}
\usepackage{hyperref}
\usepackage{url}
\usepackage{booktabs}
\usepackage{amsfonts}
\usepackage{amsmath}
\usepackage{amssymb}
\IfFileExists{nicefrac.sty}{\usepackage{nicefrac}}{}
\usepackage{microtype}
\usepackage{xcolor}
\usepackage{graphicx}
\usepackage{fontawesome5}   
\usepackage{subcaption}
\usepackage{multirow}
\usepackage{enumitem}
\usepackage{pgfplots}
\pgfplotsset{compat=1.18}
\usetikzlibrary{decorations.pathreplacing,calc}
\usepackage{algorithm}
\usepackage{algpseudocode}

\newcommand{\Kset}{\mathcal{K}}
\newcommand{\pigate}{\pi_{\mathrm{gate}}}

\usepackage{enumitem}
\setlist{itemsep=0.4\parsep,topsep=0pt,parsep=0pt,partopsep=0pt,leftmargin=1em,wide=0pt}

\definecolor{accent}{HTML}{293681}      
\colorlet{accentlight}{accent!12}
\definecolor{rule}{HTML}{B0B7C0}        
\definecolor{footbg}{HTML}{F6F8FB}      

\hypersetup{
  colorlinks=true,
  linkcolor={accent!90!black},
  citecolor={accent!90!black},
  urlcolor={accent},
  filecolor={accent}
}

\makeatletter
\renewcommand{\@maketitle}{\vbox{\hsize\textwidth
  \centering
  {\LARGE \bfseries\@title\par}
  \vskip 0.2in
  \def\And{\end{tabular}\hfil\linebreak[0]\hfil%
          \begin{tabular}[t]{c}\bf\rule{\z@}{24\p@}\ignorespaces}%
  \def\AND{\end{tabular}\hfil\linebreak[4]\hfil%
          \begin{tabular}[t]{c}\bf\rule{\z@}{24\p@}\ignorespaces}%
  \begin{tabular}[t]{c}\bf\rule{\z@}{24\p@}\@author\end{tabular}%
  \vskip 0.3in \@minus 0.1in}}
\makeatother

\title{Finding the Time to Think: \\ Learning Planning Budgets in Real-Time RL}

\author{
\href{https://aneeshers.github.io}{Aneesh Muppidi}$^{*,1,2}$\thanks{Correspondence to:
\href{mailto:aneesh.muppidi@magd.ox.ac.uk}{\texttt{aneesh.muppidi@magd.ox.ac.uk}}.
Version of \today.} \And
\href{https://firasdarwish.com}{Firas Darwish}$^{*,3}$ \And
\href{https://dylancope.com}{Dylan Cope}$^{1}$
\And
\href{https://joao.science}{João F. Henriques}$^{\dagger,2}$ \And
\href{https://www.jakobfoerster.com}{Jakob Nicolaus Foerster}$^{\dagger,1}$ \AND
{\normalfont $^{1}$\href{https://bold-lab.ai/}{British Open-ended Learning and Discovery Lab (BOLD)}, University of Oxford} \AND
{\normalfont $^{2}$\href{https://www.robots.ox.ac.uk/~vgg/}{VGG}, University of Oxford} \AND
{\normalfont $^{3}$\href{https://www.stats.ox.ac.uk/}{Department of Statistics}, University of Oxford} \AND
{\normalfont\footnotesize $^{*}$Equal contribution. \quad $^{\dagger}$Equal advising.}
}

\begin{document}

\makeatletter
\renewcommand{\thefootnote}{\fnsymbol{footnote}}
\setcounter{footnote}{2}
\makeatother

\maketitle

\makeatletter
\renewcommand{\thefootnote}{\arabic{footnote}}
\setcounter{footnote}{0}
\renewcommand{\@makefnmark}{\hbox{\textcolor{accent}{$^{\@thefnmark}$}}}
\makeatother

\begin{abstract}
Deliberating takes time. In real-time settings, that time is not free.
Standard reinforcement learning (RL) sidesteps this as the environment
waits indefinitely for the agent's decision.
Instead, we study real-time RL environments where the environment
progresses while waiting for the agent's action.
Building on prior real-time formalizations, we introduce
\emph{variable-delay real-time RL}, where the agent chooses how long
to deliberate at each decision point since the environment progresses. 
For the planning agents we use, the right delay is state-dependent, and naively planning how long to
plan can paralyze the agent.
We instead approach this setting by training a lightweight gating policy on
top of a planner to select state-dependent planning budgets.
Across real-time Pac-Man, Tetris, Snake, Speed Hex,
and Speed Go, our gating policy outperforms fixed-budget and
heuristic baselines, and transfers to a real-time setup where the environment and agent run on two different GPUs.
\end{abstract}

\begin{center}
\href{https://github.com/Aneeshers/realtime-rl-code}{\faGithub~Code}\qquad
\href{https://huggingface.co/Aneesh19/realtime-rl-checkpoints}{%
  \raisebox{-0.18\height}{\includegraphics[height=1.05em]{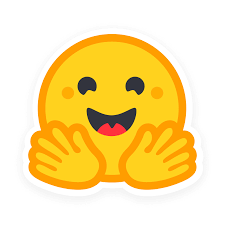}}~Checkpoints}\qquad
\href{https://aneeshers.github.io/realtime-rl/}{\faGlobe~Project Page}
\end{center}

\section{Introduction}

Expert decision-makers do not think equally hard about every choice.
They must recognize which decisions require more time or effort to
think through and which decisions can be made quickly, as they are
constrained in the cognitive resources they can devote to
decision-making \citep{otto2019opportunity}.
For example, a skilled chess player managing the clock in speed chess
will play most moves quickly and reserve their time for critical board
positions when they need to think for longer.
In standard RL, and Markov Decision Processes (MDPs), the environment
waits as the agent may deliberate indefinitely before committing to an
action, and computation does not carry any cost.

Real-time settings break this assumption: the world keeps progressing
even if the agent takes longer to act.
The cost of deliberation is not wall-clock latency, but steps taken by
the environment while the agent is still deciding.
A perfect action executed one moment too late may be far worse than a
timely, good-enough action.

\citet{ramstedt2019rtrl} formalized this concern by fixing the action
delay at exactly one timestep.
We generalize \citet{ramstedt2019rtrl}'s framework and introduce
\emph{variable-delay real-time RL}, where the agent is a procedure
that runs in time rather than a one-frame function, so the time any
particular action-producing computation spans is paid as progress in
the world.

We explore this setting with agents that use anytime planning algorithms to improve the quality of their actions, but must confront that doing so takes time in which the environment continues to evolve. Specifically, we use AlphaZero-style models \citep{silver2018general} that run Monte Carlo Tree Search (MCTS) at decision time to improve action quality (see Appendix~\ref{app:az_primer} for a primer). A property of these agents is that more search produces better actions but takes longer to run, and we verify empirically (Figure~\ref{fig:scaling}) that planning quality and inference latency scale together.

Our setting introduces three challenges.
Firstly, choosing the right delay is state-dependent: some states reward careful planning
while others demand immediate reaction.
Secondly, the agent must commit to \emph{something} during the intermediate
frames, since the environment will not wait for the planner.
Lastly, choosing how long to deliberate is fundamentally a problem of
\emph{meta-reasoning}~\citep{russell1991right,horvitz2013reasoning},
which risks an obvious \emph{paradox of planning-about-planning}: the
meta-decision occurs while the environment continues to evolve, so
deciding how long to think could incur the same per-frame cost as thinking
itself.

We address these challenges by training a lightweight \emph{gating policy} on top of a \emph{frozen}
planner that decides how long to plan at
each decision point. Our design escapes the paradox in practice because a single gate
forward pass is orders of magnitude cheaper than an MCTS rollout, and
therefore adds negligible real-time overhead. We employ our \emph{gating policy} in \emph{real-time} Pac-Man, Tetris, Snake, in which the world
progresses while the planner runs in the background (Figure~\ref{fig:realtime-rl-gating-diagram}), and \emph{clock
environments} (Speed Hex, Speed Go), in which the board is static but
each player has a clock that depletes with thinking time.
Section~\ref{sec:envs} formalizes both.

Our contributions are the following:
\begin{enumerate}[leftmargin=2em]
  \item \textbf{Variable-delay real-time RL.}
    A generalization of \citet{ramstedt2019rtrl}'s fixed-delay
    framework: under the real-time interaction protocol the agent is a
    procedure that runs in time rather than a function evaluated
    instantaneously, and the time spent producing any action is paid
    as progress in the world.
  \item \textbf{Planning quality vs.\ inference time.}
    We empirically characterize how planning quality and real-time
    inference cost co-scale with MCTS simulation count, establishing
    the joint tradeoff that motivates adaptive allocation.
  \item \textbf{Adaptive gating on a frozen planner.}
    A lightweight gating policy, trained with PPO on top of a frozen
    AlphaZero planner, that selects a state-dependent planning
    budget at each decision.
\end{enumerate}


\begingroup
\setlength{\textfloatsep}{4pt}
\begin{figure}[h!]
  \centering
  \includegraphics[width=\linewidth]{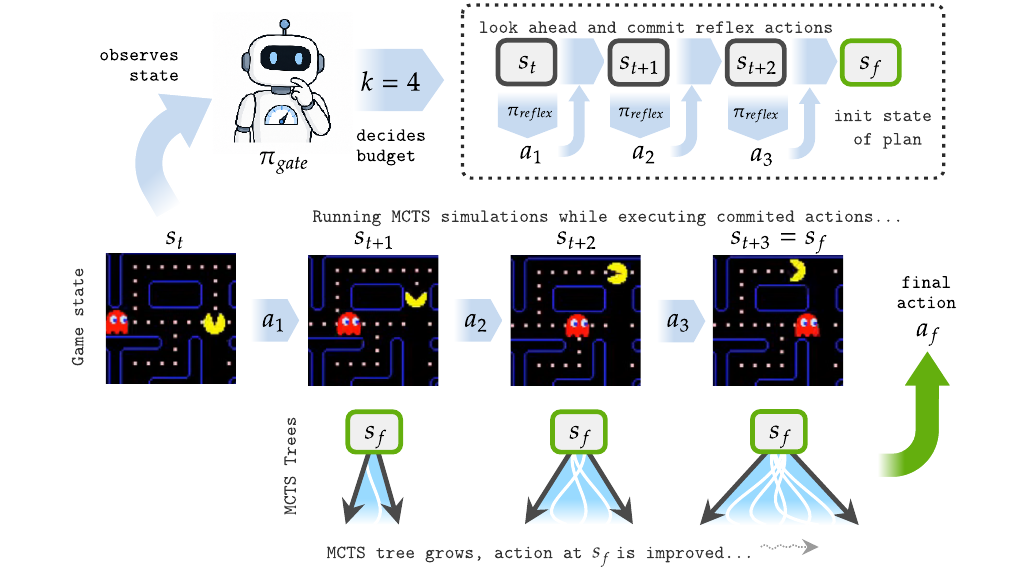}
  \caption{
    Given the current state, the gating policy chooses whether to react immediately or spend time planning, selecting
    the number of timesteps $k$ over which to plan. The agent then
    takes $k{-}1$ committed actions using $\pi_{\mathrm{reflex}}$ ($\pi_{\mathrm{0}}$) while MCTS plans,
    and finally executes the planned action.
  }
  \label{fig:realtime-rl-gating-diagram}
\end{figure}
\endgroup
Across five environments, the gating policy which adaptively selects among a set of planning budgets outperforms baselines constrained to use a single fixed planning budget at every step. To validate that the environments and method we designed capture the challenges of real-time decision-making, we transfer the learned gating policy to a real-time hardware setting where the environment runs on one GPU and the agent runs on a second GPU without any retraining or architectural changes to demonstrate that it performs effectively.




\section{Related Work}

\noindent\textbf{Real-time RL and delayed MDPs.}
The standard MDP assumes the environment waits during action selection.
\citet{travnik2018reactive} flagged this as a fundamental mismatch with
real-time interaction, and \citet{ramstedt2019rtrl} formalized an
alternative in which the agent has exactly one timestep to compute its
next action, mathematically equivalent to a 1-step constant-delay
MDP~\citep{walsh2009delayed}.
Subsequent work extends this to fixed action and observation delays of
larger length

~\citep{derman2021delayed,bouteiller2021random,katsikopoulos2003markov},
to concurrent control~\citep{xiao2020thinking}, and to asynchronous and
pipelined architectures that mitigate inference
cost~\citep{riemer2025staggered,anokhin2025handling}.
Across this literature, delay is imposed by the system.
We instead let the agent choose its delay at each decision point,
turning delay selection itself into the learning problem.

\noindent\textbf{Value of computation and meta-reasoning.}
The view that rational agents must reason not only about what to do but
about how much to think traces to bounded
rationality~\citep{simon1972theories} and Good's \emph{Type II
rationality}~\citep{good1952rational}.
\citet{russell1991right} made this operational by defining the
\emph{value of computation} (VOC) as the expected improvement in
decision quality minus its cost, and bounded
optimality~\citep{russell1991architecture} reframed rational agency as the
optimal use of fixed computational resources.
Anytime
algorithms~\citep{dean1988analysis,zilberstein1996using,likhachev2003ara,korf1990rta}
produce monotonically improving answers given more time, and Bayesian
flexible-computation methods~\citep{horvitz1989reflection} formalize
when to stop.
Applied to MCTS, value-of-computation ideas yield Bayesian formulations
of which simulation to run
next~\citep{hay2012selecting,tolpin2012simple,sezener2020static,lin2015metareasoning},
while classical chess- and Go-engine time management allocates clock
per move via hand-designed
heuristics~\citep{baier2016time,huang2010timemanagement,baudivs2011pachi}.
The cognitive-science literature models bounded deliberation itself as
a meta-MDP whose actions are
computations~\citep{lieder2017strategy,griffiths2019doing,callaway2018learning,cope_learning_2023}
and shows empirically that humans allocate planning effort to
high-stakes decisions in ways predicted by this model.
Our gating policy fits this lineage: it is a learned estimator
that chooses total budget per decision rather than which simulation to
run next, formalized as a meta-MDP over a frozen planner.

\noindent\textbf{Adaptive compute in model-based RL.}
The closest neighbors to our work learn to allocate planning compute on
top of a model-based RL agent.
Metacontrol~\citep{hamrick2017metacontrol} chooses how many imagination
steps to run; Thinker and Dynamic
Thinker~\citep{chung2023thinker,wang2024dynamic} let the agent decide
when to imagine alternative trajectories on top of a learned world
model;
\citet{hamrick2021role} provides the key empirical motivation: shallow
MuZero trees often suffice, and the marginal value of additional
simulations varies sharply by state.
AlphaZero-family
planners~\citep{silver2018general,schrittwieser2020muzero,danihelka2022policy}
treat MCTS budget as a fixed deployment hyperparameter.
Other axes of ``learning to
search''~\citep{guez2018mctsnets,farquhar2018treeqn,hamrick2020save,racaniere2017i2a}
modify the planner's internal behavior rather than its budget; we hold
the planner fixed and control only how long it runs.

\noindent\textbf{Adaptive compute in language models.}
The same thesis has been independently developed for sequence models.
PonderNet and adaptive computation
time~\citep{graves2016act,banino2021pondernet,schuster2022calm} learn
per-input halting; test-time compute
scaling~\citep{snell2024scaling,deepseek2025r1,muennighoff2025s1} and
RL-trained reasoning-length
controllers~\citep{aggarwal2025l1,muppidi2025predictive,fang2025thinkless,shen2025dast}
learn per-query thinking budgets; cascaded
routing~\citep{kim2023bigl,chen2023frugalgpt,ong2025routellm} defers
easy queries to cheap models.
The shared thesis is that a small learned policy can profitably decide
how much computation to spend per decision.
Our setting differs structurally: the cost of thinking is endogenous
to the environment dynamics, so thinking longer changes the state the
agent eventually acts in rather than incurring an external penalty.

\noindent\textbf{Our setting.}
We share the thesis of the works above but address a setting none of
them does.
We generalize \citet{ramstedt2019rtrl}'s fixed-delay framework to
variable delay, where the cost of deliberation is paid in
\emph{environmental progression} rather than artificial reward
shaping, so the gating policy learns the value of simulation budgets
from outcomes alone.
We gate a \emph{frozen} AlphaZero-style planner with no joint
training, formalized as a meta-MDP over options~\citep{sutton1999between,bacon2017optioncritic}
whose holding times are the chosen budgets.
We also introduce a protocol that makes this real-time cost legible during training and that describes the dynamics of true asynchronous execution, so the trained policy transfers to a
two-GPU deployment with no architectural change.



\section{Variable-Delay Real-Time RL}
\label{sec:envs}

\subsection{Problem: real-time MDPs}
\label{sec:problem}

Consider a standard MDP $E = (\mathcal{S}, \mathcal{A}, P, r, \gamma)$.
The \emph{real-time interaction protocol} changes only how the agent
and environment exchange actions: the environment advances by one
frame at fixed intervals $t = 0, 1, 2, \dots$ regardless of the agent,
applying at each frame whatever action the agent has submitted by
then and defaulting to an environment-defined fallback (typically a
no-op) if none.
The MDP and the return $\mathbb{E}\!\left[\sum_t \gamma^t r_t\right]$
are unchanged.

The agent is therefore no longer a function
$\pi: \mathcal{S} \to \Delta(\mathcal{A})$ but a \emph{procedure} that
runs in time and emits an action at each frame.
This is the only change, but it has a sharp consequence: thinking
costs progress in the world, since the environment advances while the
agent computes.

\paragraph{Relation to existing real-time RL.}
The Real-Time MDP of \citet{ramstedt2019rtrl} is the special case in
which the agent's procedure is constrained to be a function
$\pi: \mathcal{S} \to \Delta(\mathcal{A})$ that takes exactly one
frame to evaluate---equivalent to a 1-step constant-delay
MDP~\citep{walsh2009delayed}.
Fixed-delay MDPs of length
$K$~\citep{walsh2009delayed,derman2021delayed} fix the procedure's
latency to $K$ frames.
Our setting subsumes both: any procedure satisfying the
one-action-per-frame contract is valid, and the agent (not the
environment) decides how many frames any particular computation
spans.

\subsection{Solution: an SMDP over budgeted options}
\label{sec:smdp}

The real-time protocol leaves the agent's procedure unspecified.
Our framework constructs one from three ingredients: a fast
\emph{reflex policy} that supplies an action every frame, a finite
set of slow-but-better \emph{anytime action-refinement computations}
that the agent invokes when it can afford to wait, and a learned
\emph{gating policy} that decides at each meta-decision which
computation (if any) to run.
The first two are combined into temporally extended \emph{budgeted
options}; the gating policy operates as a meta-policy over those
options in a semi-Markov decision process (SMDP).

\paragraph{Reflex policy.}
We commit to a single fast policy $\pi_{\mathrm{reflex}}(a \mid s)$
that runs in well under one frame.
It supplies the agent's frame-by-frame output under the real-time
protocol regardless of what else is being computed in the background.
Sections~\ref{sec:committed} and~\ref{sec:clock} instantiate
$\pi_{\mathrm{reflex}}$ per environment.

\paragraph{Anytime action-refinement computations.}
We additionally equip the agent with a finite family of
\emph{anytime action-refinement computations}
$\{c_k\}_{k \in \Kset}$~\citep{russell1991right,zilberstein1996using},
indexed by discrete duration $k \in \Kset$: each $c_k$ is an anytime
algorithm whose job is to refine the action choice given more
compute.
Computation $c_k$ runs for exactly $k$ frames once initiated and, at
the start of its $k$-th frame, produces a refined action distribution
$\pi_k(\cdot \mid s_{t_n})$ at the state $s_{t_n}$ where it was
initiated; longer-running $c_k$ produce expectedly better actions, a
property we verify for our instantiation in
Figure~\ref{fig:scaling}.
Throughout this paper we instantiate $c_k$ as MCTS run for $k$
frames from $s_{t_n}$, but the framework applies to any anytime
action-refinement algorithm with known per-budget duration.

\paragraph{Budgeted options.}
For each $k \in \Kset$ we define a \emph{budgeted option}
$o_k$\footnote{Our budgeted options are a special case of the
semi-Markov options of \citet{sutton1999between}: an option is a
triple $\langle \mathcal{I}, \pi, \beta \rangle$ with input set
$\mathcal{I}$, internal policy $\pi$, and termination condition
$\beta$. \citet{sutton1999between} note that ``[s]ometimes it is
useful for options to `timeout,' to terminate after some period of
time has elapsed even if they have failed to reach any particular
state,'' and that this requires the semi-Markov generalization in
which $\beta$ may depend on the history since the option was
initiated. Each $o_k$ uses exactly this construction: deterministic
termination after $k$ primitive frames. See
\href{http://www.incompleteideas.net/papers/SPS-98.pdf}{Sutton,
Precup, and Singh (1999), \S 4} for the full formalism.}
that wraps $c_k$ into a single temporally extended action: $o_k$
initiates $c_k$, emits $k-1$ \emph{committed actions} drawn from
$\pi_{\mathrm{reflex}}$ during the wait window, and on the terminal
frame applies $c_k$'s output.
Concretely, if $o_k$ is initiated in state $s_t$, then for the
intermediate frames $j = 0, \dots, k-2$,
\begin{equation}
  a_{t+j} \sim \pi_{\mathrm{reflex}}(\cdot \mid s_{t+j}),
  \qquad
  s_{t+j+1} \sim P(\cdot \mid s_{t+j}, a_{t+j}),
\end{equation}
and on the terminal frame
\begin{equation}
  a_{t+k-1} \sim \pi_k(\cdot \mid s_t),
  \qquad
  s_{t+k} \sim P(\cdot \mid s_{t+k-1}, a_{t+k-1}),
\end{equation}
where $\pi_k$ is the action distribution produced by $c_k$ initiated
at $s_t$.
The option induces a transition kernel
$P_k(s' \mid s) = \Pr(s_{t+k} = s' \mid s_t = s, o_k)$ and an
option-level reward
\begin{equation}
  R_k(s) \;=\; \mathbb{E}\!\left[
    \sum_{j=0}^{k-1} \gamma^j \, r(s_{t+j}, a_{t+j})
    \;\Big|\; s_t = s, o_k
  \right].
\end{equation}

\begingroup
\setlength{\textfloatsep}{4pt}
\begin{figure}[h!]
  \centering
  \includegraphics[width=0.9942\linewidth]{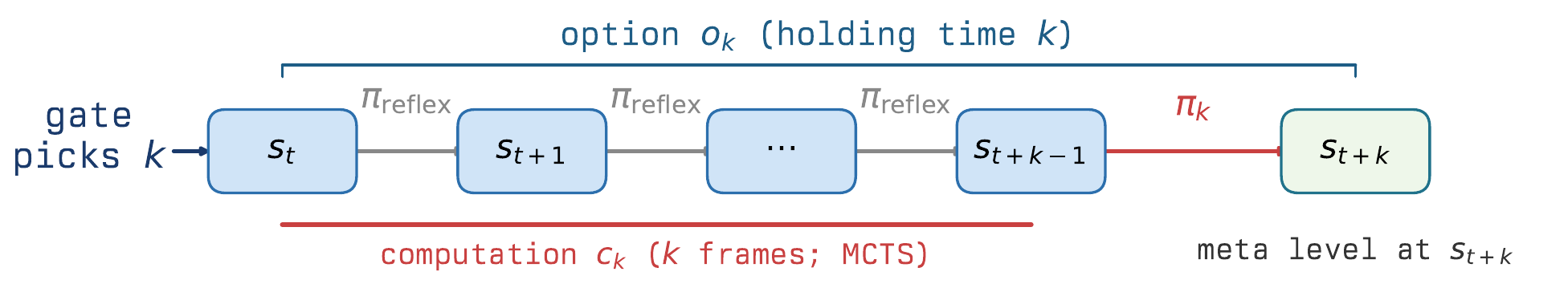}
  \vspace{-4pt}
  \caption{Timeline of a budgeted option $o_k$. The agent emits $k{-}1$
    committed actions from $\pi_{\mathrm{reflex}}$ while computation
    $c_k$ runs (instantiated as MCTS for $k$ frames), then applies
    $c_k$'s output $\pi_k$ and returns to the meta level at
    $s_{t+k}$. In clock environments, the committed steps are no-ops that
    only consume clock.}
  \vspace{-6pt}
  \label{fig:option-timeline}
\end{figure}
\endgroup

\paragraph{Meta-level SMDP.}
The gating policy $\pigate(k \mid s_t)$ is a meta-policy that chooses
among the $|\Kset|$ budgeted options.
At each meta-decision state $s_t$ the agent samples $k$, executes
$o_k$, and returns to the meta level in state $s_{t+k}$ after $k$
primitive frames.
The induced control problem is a semi-Markov decision
process~\citep{sutton1999between,bacon2017optioncritic} with
meta-state space $\mathcal{S}$, meta-action space $\Kset$, holding
time $\tau(k) = k$, kernel $P_k$, and reward $R_k$.
Its meta-Bellman equation is
\begin{equation}
  V(s_t)
  \;=\;
  \mathbb{E}_{k \sim \pigate(\cdot \mid s_t)}\!\left[\,
    \underbrace{\sum_{j=0}^{k-1} \gamma^j r_{t+j}}_{\text{option reward over $k$ frames}}
    \;+\;
    \underbrace{\gamma^k V(s_{t+k})}_{\text{discounted bootstrap}}
  \,\right].
  \label{eq:smdp}
\end{equation}
The gating policy is therefore solving a discrete control problem:
not which primitive action to take now, but which temporally
extended computation-and-control routine to invoke.
In simulation we throttle MCTS to a fixed number of simulations per
frame (32 in the committed-action environments;
see Section~\ref{sec:committed}), so the holding time of $o_k$ is
exactly $k$ frames by construction and this SMDP has deterministic
holding times.
In real-time deployment, MCTS runtime is slightly stochastic, so the
hardware system is closer to a continuous-time SMDP.
Appendix~\ref{app:rtrl_smdp} discusses this distinction.

\paragraph{Why options rather than state augmentation.}
The RTMDP framework~\citep{ramstedt2019rtrl} handles its fixed
one-step delay by augmenting the state with the action currently in
flight, $\mathbf{x}_t = (s_t, a_t)$, with transition kernel
\begin{equation}
  P_{\mathrm{RT}}(\mathbf{x}_{t+1} \mid \mathbf{x}_t, \hat{a}_t)
  = P(s_{t+1} \mid s_t, a_t)\,\delta(a_{t+1} - \hat{a}_t),
\end{equation}
where $\delta$ is a Dirac delta.
This is the natural representation when delay is fixed and the agent
emits nothing during the wait window.
In our setting both assumptions fail: $k$ is chosen by the agent at
every decision frame, and during the wait window the agent is actively
controlling the environment through $\pi_{\mathrm{reflex}}$.
State augmentation would have to grow with $|\Kset|$ and encode the
within-option phase; budgeted options collapse the within-option
dynamics into a single SMDP transition, leaving the gating policy a
standard meta-decision problem.

\subsection{Committed-action environments: Pac-Man, real-time Tetris, Snake}
\label{sec:committed}

In committed-action environments, acting and planning happen
concurrently.
The $k-1$ committed actions are drawn from $\pi_{\mathrm{reflex}}$,
which we instantiate as the planner's own policy network evaluated
without MCTS, computed via a single forward pass in roughly two
milliseconds, while the full MCTS runs in the background.
We choose this instantiation rather than a separately trained reflex
policy because it requires no additional training and keeps the
$\pi_{\mathrm{reflex}}$ aligned with the MCTS tree's internal rollouts.
On the $k$-th frame the planner's chosen action is applied. MCTS rolls out the same $k{-}1$ committed steps that the
agent will actually execute, so search is performed over the future
state in which the planned action will land.
The cost of deeper planning is therefore exactly $k{-}1$ additional
frames of world progression before the carefully chosen action
lands.

We instantiate this setting in three environments derived from
\texttt{Jumanji} ~\citep{bonnet2024jumanji}: Pac-Man, Snake, and a real-time
variant of Jumanji Tetris (Appendix~\ref{app:envs_tetris}).
The source of progression differs across environments (ghost motion
in Pac-Man, gravity in real-time Tetris, body elongation in Snake),
but in all cases a deeper plan arrives later, after the world has
changed.

We use $\Kset = \{1, 2, 3, 4\}$ and calibrate 32 MCTS simulations to
one environment frame, which corresponds to roughly 9\,FPS on an H100
(Section~\ref{sec:deployment}).
At budget $k$ the planner runs $32k$ simulations over $k$
frames, so each frame still receives 32 simulations on average
regardless of $k$.
Differences in performance reflect \emph{when} compute is allocated,
not the average compute spent per frame.

\subsection{Clock environments: Speed Hex and Speed Go}
\label{sec:clock}

Clock environments instantiate our framework with a degenerate reflex
policy: $\pi_{\mathrm{reflex}}$ deterministically emits the no-op
action, so committed steps leave the board unchanged and only $c_k$'s
terminal action affects play.
The clock is therefore the only dynamic element of the environment;
it decrements with thinking time over the $k$ frames of an option,
while the board state remains fixed until the planner's chosen
action is applied.
Intuitively, this is the same pressure as in speed chess: the board
waits for your move, but your limited clock keeps running while you
think.

We instantiate this setting in Speed Hex ($11 \times 11$) and Speed
Go ($9 \times 9$), both board-game environments built on \texttt{pgx} \citep{koyamada2023pgx}.
We use a one-to-one mapping between MCTS simulations and clock
ticks: each simulation increments the player's clock by one unit.

For clock environments the equal-compute-per-frame constraint does
not apply, so we calibrate the simulation options separately to
ensure each option represents a meaningfully distinct tradeoff
between inference latency and planning quality.
Calibration details and the resulting option sets
$\Kset = \{2, 8, 32, 128\}$ for Speed Hex and
$\Kset = \{16, 32, 64, 96\}$ for Speed Go are in
Appendix~\ref{app:sim_calibration}.
\section{Adaptive Gating Policy}
\label{sec:method}

\subsection{The planning tradeoff}

The core observation motivating our approach is that planning quality
and inference cost scale together.
As an agent runs more MCTS simulations, its action quality improves;
but more simulations also take longer to run, and in real-time
settings that longer inference is exactly what advances the
environment, or depletes the clock, before the agent acts.
We empirically verify this relationship in Figure~\ref{fig:scaling}.
The full five-environment scaling results appear in Appendix~\ref{app:sim_calibration}. This joint scaling means
we do not need to learn \emph{how} to plan as the planner handles
that, only \emph{when} the additional quality gain is worth the
additional real-time cost.

\subsection{The gating policy}

The gating policy takes three inputs at each meta-step: (1) the raw
game observation, (2) the planner's intermediate spatial features extracted
from its frozen trunk and (3) the planner's scalar value
estimate $V(s_t)$.
In clock environments the observation also includes the remaining
time budget.
Together these give the gating policy both a direct view of the board
and a compressed summary of the planner's own confidence, without
requiring any access to the planner's internals beyond a single
forward pass.
A lightweight network processes these inputs and produces a
distribution over $k$ and a baseline value for training.
All architecture details are in Appendix~\ref{app:implementation}.
\begin{figure}[t]
  \centering
  \includegraphics[width=\linewidth]{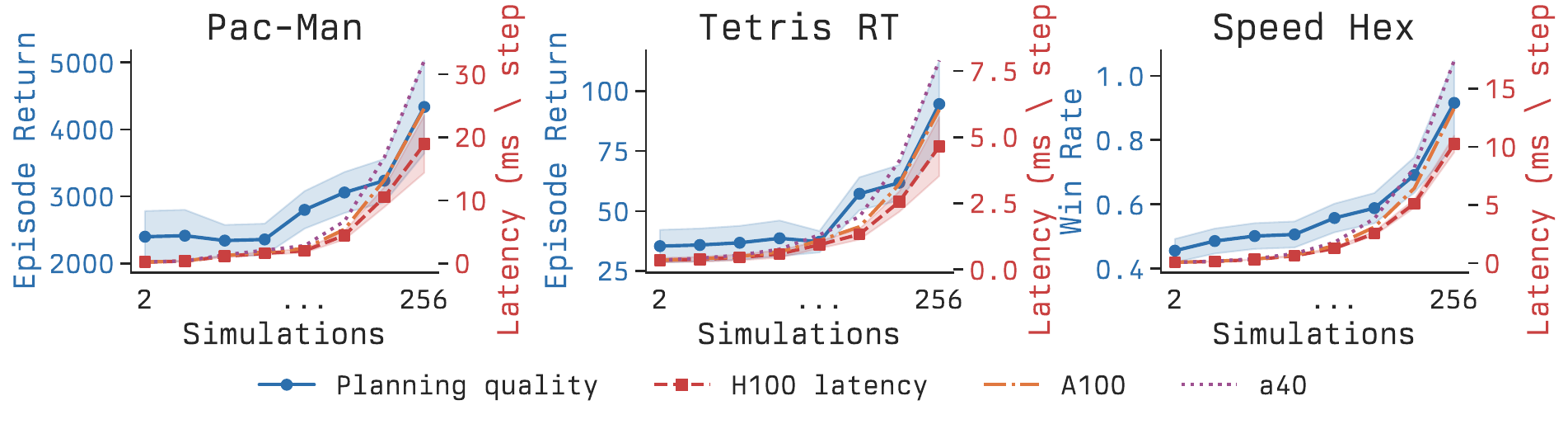}
  \caption{
    Across Pac-Man, Tetris, and 2-player Speed Hex, planning
    quality rises with simulation count while inference latency rises
    alongside it.
    Blue denotes performance; dashed curves denote per-step latency on
    H100, A100, and A40; shading shows $\pm$SE.}
  \label{fig:scaling}
\end{figure}
\vspace{-9pt}
\subsection{Training}

We train in two phases.
First, we train AlphaZero-style base planners for each environment.
We then freeze the selected base planner and train the gating policy
with PPO~\citep{schulman2017proximal} on top of it.
For the clock environments, these base planners are trained by
self-play (Appendix~\ref{app:implementation}).
Because each meta-step has variable duration $k$, we adapt
Generalized Advantage Estimation~\citep{schulman2015high} to carry
the per-meta-step discount $\gamma^k$ through the advantage
computation (see Appendix~\ref{app:gae}).
Making this meta-MCTS AlphaZero stack computationally feasible
requires careful attention to several JAX implementation details,
which we summarize in Appendix~\ref{app:implementation}.

\section{Experiments}
\label{sec:experiments}

\begin{figure}[t]
  \centering
  \includegraphics[width=\linewidth]{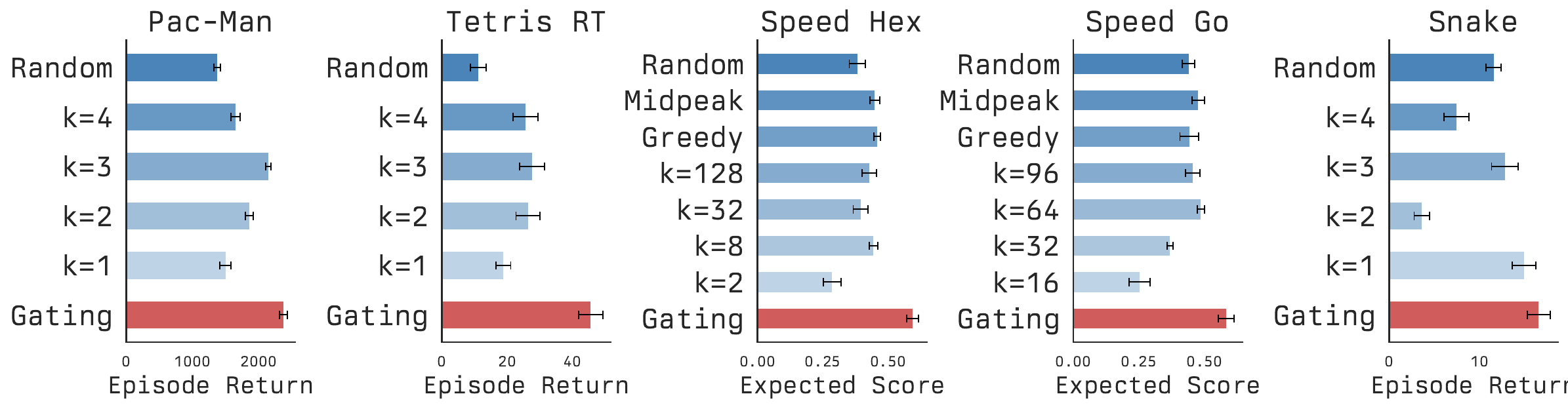}
  \caption{
    Across all five environments, the gating policy outperforms fixed-budget and
    heuristic baselines, showing that adaptive allocation matters more than
    committing to a single search budget. Bars show mean $\pm$\,SE over 100
    episodes; for Speed Hex and Speed Go, expected score is averaged over the
    shared sampled clock budgets.}
  \label{fig:main_results}
\end{figure}

\paragraph{Baselines.}
For real-time Pacman, Tetris, and Snake we compare against: \emph{always-$k$} policies for
each $k \in \Kset$; and a \emph{random} policy selecting $k$ uniformly at each
meta-step.
For speed environments we additionally compare against \emph{greedy} (always use
maximum simulations) and \emph{midpeak} (a bell-curve allocation that concentrates
compute at the critical mid-game, a common heuristic in game engines \cite{baier2016time, huang2010timemanagement}). For these clock heuristics we tune the bell-curve shape separately at each evaluation budget and report the best resulting strategy. All results are averaged over 100 independent episode seeds; we report mean $\pm$ standard error in Fig.~\ref{fig:main_results}.



\vspace{-6pt}

For both Speed Hex and Speed Go, we evaluate across the same five sampled clock
budgets, $T \in \{300, 1200, 2300, 3500, 4100\}$, and average the resulting
head-to-head expected scores. The fixed-budget baselines use 2/8/32/128
simulations for Speed Hex and 16/32/64/96 simulations for Speed Go.
In these main clock benchmarks, exhausting the clock does not immediately end the
game. We found that simply exhausting the clock was too easy a setting
(Appendix~\ref{app:hex_timeout_control}), so we study the harder
$\pi_{\mathrm{reflex}}$-fallback setting, where control passes to
$\pi_{\mathrm{reflex}}$ after the clock runs out and the task becomes
one of resource allocation rather than simply avoiding timeout.

\paragraph{The gating policy outperforms every fixed-budget baseline across all environments.} In Pac-Man, the best fixed policy ($k{=}3$, 96 sims/step) scores 2149; the gating policy reaches 2370 ($+10.3\%$). In real-time Tetris, the best fixed policy ($k{=}3$) scores 27.6; the gating policy reaches 45.6 ($+65\%$).  In Snake, the best fixed policy ($k{=}1$) scores 14.91; the gating policy reaches 16.54 ($+10.9\%$). In Speed Hex, averaged over the shared sampled clock budgets, the gating policy reaches an expected score of 0.58, compared with 0.43 for the best fixed-budget baseline ($k{=}128$) and 0.46 for the best heuristic baseline (greedy). In Speed Go, averaged over the same budgets, the gating policy reaches an expected score of 0.59, compared with 0.51 for the best fixed-budget baseline ($k{=}64$) and 0.50 for the best heuristic baseline (midpeak).

Across environments, the gate learns dynamic, state and budget-dependent allocation strategies that outperform both fixed-budget policies and hand-designed heuristics (Figure~\ref{fig:strategy}, Appendix Figure~\ref{fig:strategy_appendix}).
The random baseline falls well below every fixed-$k$ policy in both environments, confirming that \emph{when} to allocate compute matters as much as how much.

\section{Analysis}
\label{sec:discussion}

\subsection{What triggers deeper planning?}
\label{subsec:whattriggersdeeperplanning}

\begin{figure}[h]
  \centering
  \includegraphics[width=\linewidth]{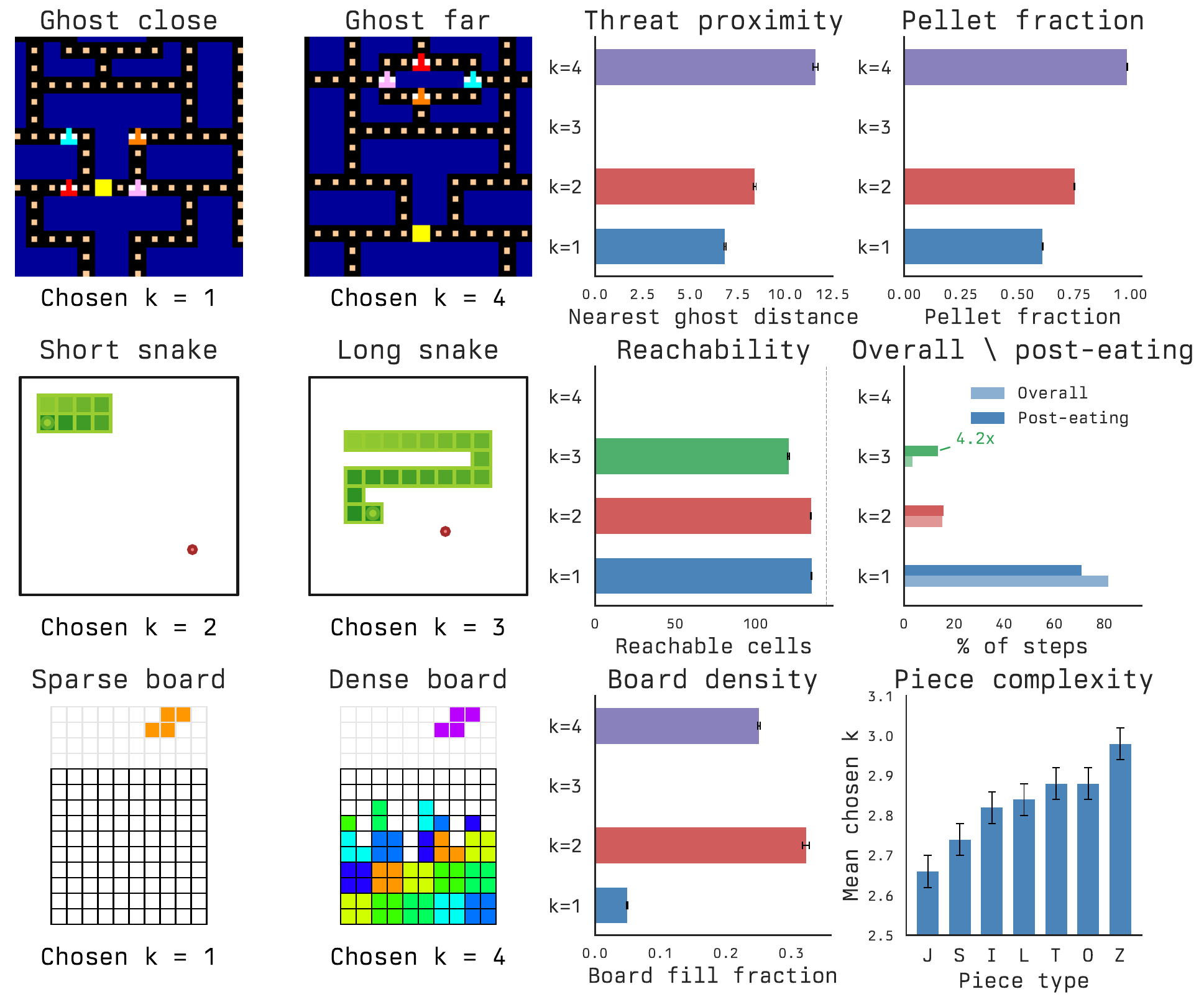}
  \caption{The policy plans deeply precisely when the state is dangerous or constrained.
    Across Pac-Man, real-time Tetris, and Snake, larger chosen budgets are associated with
    higher threat, denser boards, or fewer safe continuations, indicating that
    the gate is responding to meaningful decision difficulty. Plots show state
    features conditioned on chosen budget $k$ (mean $\pm$\,1\,SE, 100 episodes).}
\label{fig:interpretability}
\end{figure}

Across the environments, our gating policy allocates compute in
states where the consequence of a suboptimal action is large and additional search
budget is likely to change the decision (Figure~\ref{fig:interpretability}).

\paragraph{Pac-Man.}

In the current evaluation set, larger chosen budgets are associated with greater
nearest-ghost distance in Pac-Man: when the nearest ghost is close, the policy
defaults to the more reactive $k{=}1$ option; at intermediate distances it shifts
toward $k{=}2$; and when the threat is farthest away it can afford the deeper
$k{=}4$ plan. We also see in Appendix~\ref{app:strategy_appendix} that the
gating policy becomes more reactive late in the episode.

\paragraph{Real-time Tetris.}

Board density is the dominant feature in real-time Tetris.
$k{=}1$ is selected almost exclusively on near-empty boards
(fill $\approx 0.05$, stack $\approx 3.5$ rows) where any reasonable placement is
acceptable.
$k{=}2$ and $k{=}4$ are selected on dense boards (fill $\approx 0.25$--$0.32$,
stacks $\approx 12$ rows) where placement precision is consequential.
$k{=}3$ is never used ($0\%$), producing the bimodal ``react or plan deeply'' strategy
visible in Figure~\ref{fig:interpretability}.
Piece type also modulates budget: the Z-piece ($\bar{k}{=}2.98{\scriptstyle\pm0.04}$)
and O- and T-pieces ($\bar{k}{=}2.88{\scriptstyle\pm0.04}$) receive the most
deliberation, while the J-piece ($\bar{k}{=}2.66{\scriptstyle\pm0.04}$) receives the
least, reflecting differences in placement geometry and expected outcome variance.

\paragraph{Snake.}

Spatial constraint, rather than simple goal distance, is the main trigger.
$k{=}1$ dominates on open boards, while $k{=}3$ appears when reachability drops,
local body density rises, and the snake has fewer safe continuations. $k{=}2$
occupies a milder regime associated with navigating toward more distant fruit on
otherwise open boards. Immediately after eating, when the body has just grown,
$k{=}3$ usage jumps from $3.2\%$ to $13.5\%$, indicating that the policy treats
body-growth events as a prospective spatial-risk signal.

\paragraph{Clock environments.}

\begin{figure}[h]
  \centering
  \includegraphics[width=1\linewidth]{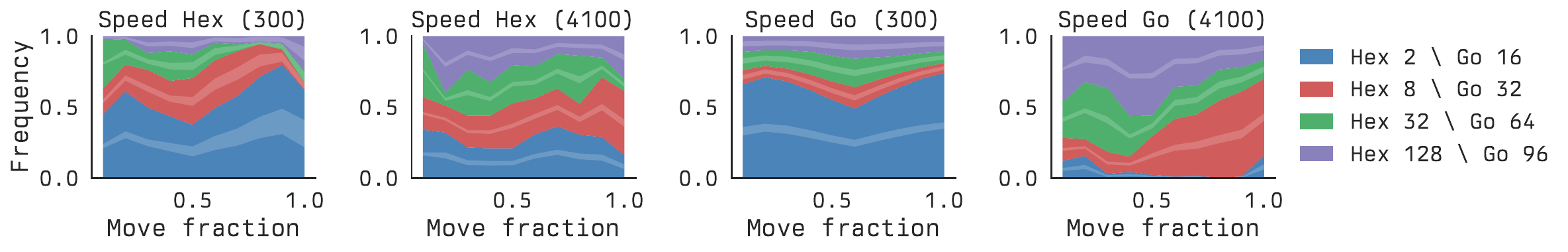}
  \caption{ In both Speed Hex and
    Speed Go, the learned policy is much more reactive under the small budget
    ($T{=}300$) and distributes mass toward deeper options once more clock is
    available ($T{=}4100$).}
  \label{fig:strategy}
\end{figure}

In the clocked PGX games, a key trigger is
the interaction between board state and remaining time. Figure~\ref{fig:strategy}
shows the resulting shift in allocation for both Speed Hex and Speed Go. Under the
small clock budget ($T{=}300$), the policy heavily favors the cheapest option,
reflecting the high opportunity cost of thinking. Under the larger clock budget
($T{=}4100$), the same learned policy spreads probability mass much more evenly
across the budgets, allocating deeper search substantially more often. Because
the policy is trained across many clock settings rather than tuned separately at
each one, this shift is evidence of budget-conditioned generalization.

\section{Real-Time Deployment}
\label{sec:deployment}

\begin{figure}[h!]
  \centering
  \includegraphics[width=\linewidth]{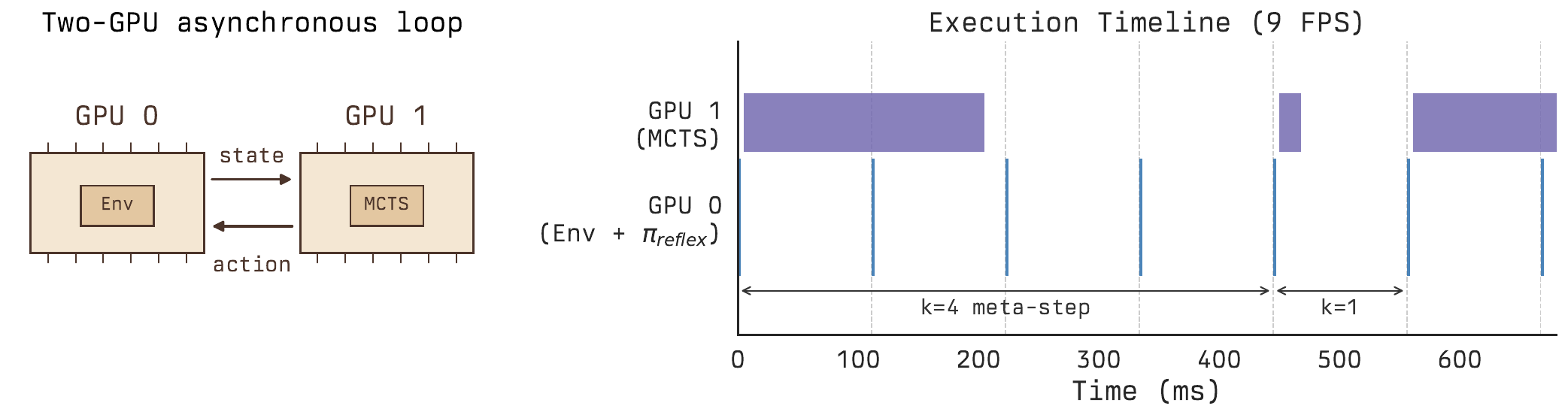}
  \caption{\textbf{Two-GPU asynchronous deployment pipeline.}
    GPU~0 runs the environment continuously via committed (reflex) actions;
    GPU~1 runs MCTS in parallel, so a larger $k$ buys more planning time without
    pausing the game.
    \emph{Left:} schematic of the two-GPU loop.
    \emph{Right:} execution trace at 9\,FPS, a $k{=}4$ meta-step spans four 111\,ms
    frames, followed by a $k{=}1$ recovery step.}
  \label{fig:deployment_timeline}
\end{figure}

We validate the committed-action framework in a two-GPU asynchronous setup across
three committed-action environments: real-time Tetris, Pac-Man, and Snake.
One GPU runs the environment and executes committed actions continuously; a second GPU
runs the MCTS planner.
At each meta-decision the current game state is transferred to the planning GPU, which
begins search while the environment GPU keeps the game moving with committed actions.
When planning finishes, after $k \times 32$ simulations, the result is applied at
the next environment step (Figure~\ref{fig:deployment_timeline}).

\begin{figure}[h!]
  \centering
  \includegraphics[width=\linewidth]{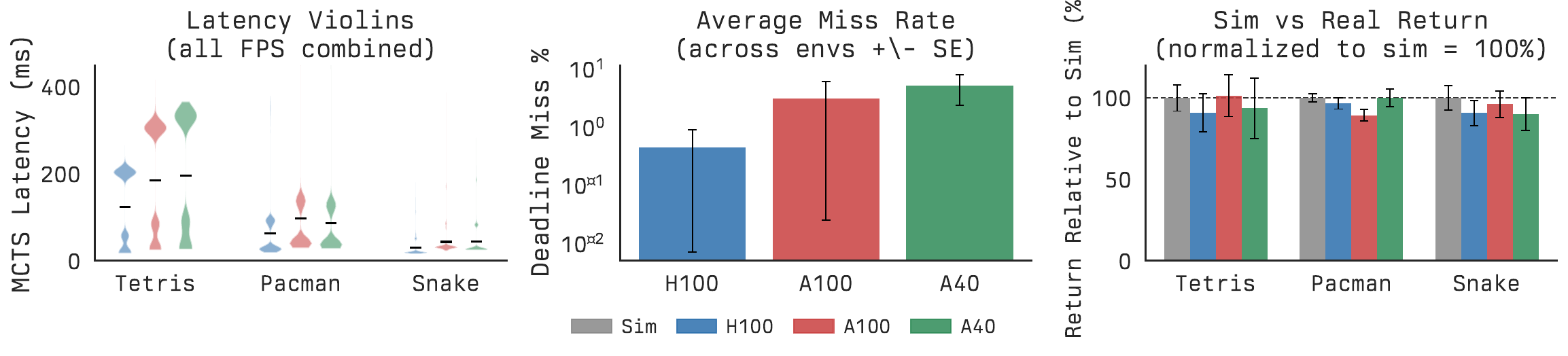}
  \caption{\textbf{Simulation-trained policies transfer cleanly to hardware deployment.}
    \emph{Left:} MCTS latency variance tightens with GPU class---H100 is most
    consistent, A100 intermediate, A40 widest.
    \emph{Centre:} H100 and A100 miss deadlines rarely; A40 breaks down only at
    the tightest frame rates.
    \emph{Right:} Deployed returns at 9\,FPS match simulation closely across all
    environments and GPU classes.}
  \label{fig:deployment}
\end{figure}

Figure~\ref{fig:deployment} summarizes 45 measured deployments:
3 environments $\times$ 3 GPU classes $\times$ 5 frame rates (8--12\,FPS).
The asynchronous execution pattern is unchanged from training (see
Section~\ref{sec:committed}): a $k{=}4$
meta-step at 9\,FPS spans four 111\,ms frames, while committed actions from the reflex policy keep the
environment moving and MCTS finishes asynchronously on the second GPU.

The main result, as seen in Fig.~\ref{fig:deployment}, is that simulation-trained policies transfer cleanly to asynchronous
hardware deployment over a broad operating range.
Latency remains well within budget on H100 in all three environments, and remains
reliable on A100 except near the tightest deadline regimes. Return transfer is similarly strong.
At 9\,FPS, deployed returns remain close to their simulation counterparts across all
three environments, and the learned budget distribution also transfers cleanly.
In real-time Tetris on H100, the policy preserves the same strongly bimodal
``react or plan deeply'' strategy seen in simulation.
Appendix~\ref{app:deployment_details} gives the full per-environment, per-FPS,
per-GPU analysis.

This result tests a key hypothesis behind the committed-action training protocol.
We treat the $k{-}1$ committed steps executed in simulation as a model of real
asynchronous deployment: if this abstraction is correct, then a policy trained
under it should transfer to a hardware-separated environment-and-planner system
without modification. Because training simulates the planning delay by actually
executing committed actions in the environment, the transfer result supports
this hypothesis and suggests that the separation between environment and planner
is the right abstraction for real-time deployment.

\section{Conclusion}

We generalized \citet{ramstedt2019rtrl}'s fixed-delay real-time RL
framework by treating the agent as a procedure that runs in time
rather than a function evaluated instantaneously: under the real-time
interaction protocol the environment advances every frame regardless,
and the time spent producing any action is paid as progress in the
world.
We then instantiated this with an SMDP over budgeted options, in
which a lightweight gating policy trained with PPO on top of a frozen
AlphaZero planner learns to invest more search where it matters and
react quickly elsewhere.
Across five environments spanning committed-action and clock-based
real-time mechanisms, the gating policy outperforms every fixed-budget
baseline, and the simulation-trained policy transfers to a true
two-GPU asynchronous deployment with no architectural change.
While we instantiate the framework with MCTS, it applies more broadly
to any anytime action-refinement algorithm with known per-budget
duration.

\paragraph{Limitations.}
Our committed-action protocol relies on the MCTS tree faithfully
simulating the $k{-}1$ committed steps, which assumes a perfect
environment simulator; extending to learned-dynamics planners
(e.g., MuZero) is natural future work.
The gating policy is trained on top of a \emph{frozen} base planner,
and Appendix~\ref{app:cross_eval} shows that the choice of base
checkpoint substantially affects downstream performance; joint
optimization of planner and gate remains open.
We hand-calibrate a small discrete budget set per environment;
continuous or richer compute vocabularies are natural extensions but
require recalibrating the cost-quality tradeoff.

\section*{Acknowledgments}
We thank
\href{https://justinsvegliato.com}{Justin Svegliato} for valuable
feedback on our metareasoning definitions and framing, and
\href{https://scholar.google.com/citations?user=7TVJf1gAAAAJ&hl=en}{Mattie Fellows}
and \href{https://uljad.com}{Uljad Berdica} for helpful discussions and
feedback on earlier drafts.
A.~Muppidi and F.~Darwish are supported by the
\href{https://www.rhodeshouse.ox.ac.uk}{Rhodes Scholarship} (Rhodes
Trust). The authors declare no competing interests.


\bibliographystyle{rlc}
\bibliography{refs}

@article{ramstedt2019rtrl,
  title={Real-time reinforcement learning},
  author={Ramstedt, Simon and Pal, Chris},
  journal={Advances in neural information processing systems},
  volume={32},
  year={2019}
}

@article{travnik2018reactive,
  title={Reactive reinforcement learning in asynchronous environments},
  author={Travnik, Jaden B and Mathewson, Kory W and Sutton, Richard S and Pilarski, Patrick M},
  journal={Frontiers in Robotics and AI},
  volume={5},
  pages={79},
  year={2018},
  publisher={Frontiers Media SA}
}

@article{walsh2009delayed,
  title={Learning and planning in environments with delayed feedback},
  author={Walsh, Thomas J and Nouri, Ali and Li, Lihong and Littman, Michael L},
  journal={Autonomous Agents and Multi-Agent Systems},
  volume={18},
  number={1},
  pages={83--105},
  year={2009},
  publisher={Springer}
}

@article{derman2021delayed,
  title={Acting in delayed environments with non-stationary markov policies},
  author={Derman, Esther and Dalal, Gal and Mannor, Shie},
  journal={arXiv preprint arXiv:2101.11992},
  year={2021}
}

@inproceedings{bouteiller2021random,
  title={Reinforcement learning with random delays},
  author={Bouteiller, Yann and Ramstedt, Simon and Beltrame, Giovanni and Pal, Christopher and Binas, Jonathan},
  booktitle={International conference on learning representations},
  year={2020}
}

@article{katsikopoulos2003markov,
  title={Markov decision processes with delays and asynchronous cost collection},
  author={Katsikopoulos, Konstantinos V and Engelbrecht, Sascha E},
  journal={IEEE transactions on automatic control},
  volume={48},
  number={4},
  pages={568--574},
  year={2003},
  publisher={IEEE}
}

@article{xiao2020thinking,
  title={Thinking while moving: Deep reinforcement learning with concurrent control},
  author={Xiao, Ted and Jang, Eric and Kalashnikov, Dmitry and Levine, Sergey and Ibarz, Julian and Hausman, Karol and Herzog, Alexander},
  journal={arXiv preprint arXiv:2004.06089},
  year={2020}
}

@article{riemer2025staggered,
  title={Enabling realtime reinforcement learning at scale with staggered asynchronous inference},
  author={Riemer, Matthew and Subbaraj, Gopeshh and Berseth, Glen and Rish, Irina},
  journal={arXiv preprint arXiv:2412.14355},
  year={2024}
}

@article{anokhin2025handling,
  title={Handling Delay in Real-Time Reinforcement Learning},
  author={Anokhin, Ivan and Rishav, Rishav and Riemer, Matthew and Chung, Stephen and Rish, Irina and Kahou, Samira Ebrahimi},
  journal={arXiv preprint arXiv:2503.23478},
  year={2025}
}

@article{silver2018general,
  title={A general reinforcement learning algorithm that masters chess, shogi, and Go through self-play},
  author={Silver, David and Hubert, Thomas and Schrittwieser, Julian and Antonoglou, Ioannis and Lai, Matthew and Guez, Arthur and Lanctot, Marc and Sifre, Laurent and Kumaran, Dharshan and Graepel, Thore and others},
  journal={Science},
  volume={362},
  number={6419},
  pages={1140--1144},
  year={2018},
  publisher={American Association for the Advancement of Science}
}

@article{schrittwieser2020muzero,
  title={Mastering atari, go, chess and shogi by planning with a learned model},
  author={Schrittwieser, Julian and Antonoglou, Ioannis and Hubert, Thomas and Simonyan, Karen and Sifre, Laurent and Schmitt, Simon and Guez, Arthur and Lockhart, Edward and Hassabis, Demis and Graepel, Thore and others},
  journal={Nature},
  volume={588},
  number={7839},
  pages={604--609},
  year={2020},
  publisher={Nature Publishing Group UK London}
}

@inproceedings{danihelka2022policy,
  title={Policy improvement by planning with Gumbel},
  author={Danihelka, Ivo and Guez, Arthur and Schrittwieser, Julian and Silver, David},
  booktitle={International Conference on Learning Representations},
  year={2022}
}

@article{hamrick2017metacontrol,
  title={Metacontrol for adaptive imagination-based optimization},
  author={Hamrick, Jessica B and Ballard, Andrew J and Pascanu, Razvan and Vinyals, Oriol and Heess, Nicolas and Battaglia, Peter W},
  journal={arXiv preprint arXiv:1705.02670},
  year={2017}
}

@article{chung2023thinker,
  title={Thinker: Learning to plan and act},
  author={Chung, Stephen and Anokhin, Ivan and Krueger, David},
  journal={Advances in Neural Information Processing Systems},
  volume={36},
  pages={22896--22933},
  year={2023}
}

@inproceedings{
wang2024dynamic,
title={Dynamic Thinker: Optimizing Decision-Time Planning with Costly Compute},
author={Kevin A. Wang and Jerry Xia and Stephen Chung and Amy Greenwald},
booktitle={The Seventeenth Workshop on Adaptive and Learning Agents},
year={2025},
url={https://openreview.net/forum?id=yGglBJ1pjZ}
}

@article{hamrick2021role,
  title={On the role of planning in model-based deep reinforcement learning},
  author={Hamrick, Jessica B and Friesen, Abram L and Behbahani, Feryal and Guez, Arthur and Viola, Fabio and Witherspoon, Sims and Anthony, Thomas and Buesing, Lars and Veli{\v{c}}kovi{\'c}, Petar and Weber, Th{\'e}ophane},
  journal={arXiv preprint arXiv:2011.04021},
  year={2020}
}

@inproceedings{guez2018mctsnets,
  title={Learning to search with mctsnets},
  author={Guez, Arthur and Weber, Th{\'e}ophane and Antonoglou, Ioannis and Simonyan, Karen and Vinyals, Oriol and Wierstra, Daan and Munos, R{\'e}mi and Silver, David},
  booktitle={International conference on machine learning},
  pages={1822--1831},
  year={2018},
  organization={PMLR}
}

@article{farquhar2018treeqn,
  title={Treeqn and atreec: Differentiable tree-structured models for deep reinforcement learning},
  author={Farquhar, Gregory and Rockt{\"a}schel, Tim and Igl, Maximilian and Whiteson, Shimon},
  journal={arXiv preprint arXiv:1710.11417},
  year={2017}
}

@article{hamrick2020save,
  title={Combining q-learning and search with amortized value estimates},
  author={Hamrick, Jessica B and Bapst, Victor and Sanchez-Gonzalez, Alvaro and Pfaff, Tobias and Weber, Theophane and Buesing, Lars and Battaglia, Peter W},
  journal={arXiv preprint arXiv:1912.02807},
  year={2019}
}

@article{racaniere2017i2a,
  title={Imagination-augmented agents for deep reinforcement learning},
  author={Racani{\`e}re, S{\'e}bastien and Weber, Th{\'e}ophane and Reichert, David and Buesing, Lars and Guez, Arthur and Jimenez Rezende, Danilo and Puigdom{\`e}nech Badia, Adri{\`a} and Vinyals, Oriol and Heess, Nicolas and Li, Yujia and others},
  journal={Advances in neural information processing systems},
  volume={30},
  year={2017}
}

@article{russell1991right,
  title={Principles of metareasoning},
  author={Russell, Stuart and Wefald, Eric},
  journal={Artificial intelligence},
  volume={49},
  number={1-3},
  pages={361--395},
  year={1991},
  publisher={Elsevier}
}

@article{hay2012selecting,
  title={Selecting computations: Theory and applications},
  author={Hay, Nicholas and Russell, Stuart and Tolpin, David and Shimony, Solomon Eyal},
  journal={arXiv preprint arXiv:1408.2048},
  year={2014}
}

@inproceedings{tolpin2012simple,
  title={MCTS based on simple regret},
  author={Tolpin, David and Shimony, Solomon},
  booktitle={Proceedings of the AAAI Conference on Artificial Intelligence},
  volume={26},
  number={1},
  pages={570--576},
  year={2012}
}

@inproceedings{sezener2020static,
  title={Static and dynamic values of computation in mcts},
  author={Sezener, Eren and Dayan, Peter},
  booktitle={Conference on Uncertainty in Artificial Intelligence},
  pages={31--40},
  year={2020},
  organization={PMLR}
}

@article{lin2015metareasoning,
  title={Metareasoning for planning under uncertainty},
  author={Lin, Christopher H and Kolobov, Andrey and Kamar, Ece and Horvitz, Eric},
  journal={arXiv preprint arXiv:1505.00399},
  year={2015}
}

@article{lieder2017strategy,
  title={Strategy selection as rational metareasoning.},
  author={Lieder, Falk and Griffiths, Thomas L},
  journal={Psychological review},
  volume={124},
  number={6},
  pages={762},
  year={2017},
  publisher={American Psychological Association}
}

@article{griffiths2019doing,
  title={Doing more with less: meta-reasoning and meta-learning in humans and machines},
  author={Griffiths, Thomas L and Callaway, Frederick and Chang, Michael B and Grant, Erin and Krueger, Paul M and Lieder, Falk},
  journal={Current Opinion in Behavioral Sciences},
  volume={29},
  pages={24--30},
  year={2019},
  publisher={Elsevier}
}

@article{callaway2018learning,
  title={Learning to select computations},
  author={Callaway, Frederick and Gul, Sayan and Krueger, Paul M and Griffiths, Thomas L and Lieder, Falk},
  journal={arXiv preprint arXiv:1711.06892},
  year={2017}
}

@article{graves2016act,
  title={Adaptive computation time for recurrent neural networks},
  author={Graves, Alex},
  journal={arXiv preprint arXiv:1603.08983},
  year={2016}
}

@article{banino2021pondernet,
  title={Pondernet: Learning to ponder},
  author={Banino, Andrea and Balaguer, Jan and Blundell, Charles},
  journal={arXiv preprint arXiv:2107.05407},
  year={2021}
}

@article{schuster2022calm,
  title={Confident adaptive language modeling},
  author={Schuster, Tal and Fisch, Adam and Gupta, Jai and Dehghani, Mostafa and Bahri, Dara and Tran, Vinh and Tay, Yi and Metzler, Donald},
  journal={Advances in Neural Information Processing Systems},
  volume={35},
  pages={17456--17472},
  year={2022}
}

@article{snell2024scaling,
  title={Scaling llm test-time compute optimally can be more effective than scaling model parameters},
  author={Snell, Charlie and Lee, Jaehoon and Xu, Kelvin and Kumar, Aviral},
  journal={arXiv preprint arXiv:2408.03314},
  year={2024}
}

@article{deepseek2025r1,
  title={Deepseek-r1: Incentivizing reasoning capability in llms via reinforcement learning},
  author={Guo, Daya and Yang, Dejian and Zhang, Haowei and Song, Junxiao and Wang, Peiyi and Zhu, Qihao and Xu, Runxin and Zhang, Ruoyu and Ma, Shirong and Bi, Xiao and others},
  journal={arXiv preprint arXiv:2501.12948},
  year={2025}
}

@inproceedings{muennighoff2025s1,
  title={s1: Simple test-time scaling},
  author={Muennighoff, Niklas and Yang, Zitong and Shi, Weijia and Li, Xiang Lisa and Fei-Fei, Li and Hajishirzi, Hannaneh and Zettlemoyer, Luke and Liang, Percy and Cand{\`e}s, Emmanuel and Hashimoto, Tatsunori B},
  booktitle={Proceedings of the 2025 Conference on Empirical Methods in Natural Language Processing},
  pages={20286--20332},
  year={2025}
}

@article{aggarwal2025l1,
  title={L1: Controlling how long a reasoning model thinks with reinforcement learning},
  author={Aggarwal, Pranjal and Welleck, Sean},
  journal={arXiv preprint arXiv:2503.04697},
  year={2025}
}

@article{fang2025thinkless,
  title={Thinkless: Llm learns when to think},
  author={Fang, Gongfan and Ma, Xinyin and Wang, Xinchao},
  journal={arXiv preprint arXiv:2505.13379},
  year={2025}
}

@inproceedings{shen2025dast,
  title={Dast: Difficulty-adaptive slow-thinking for large reasoning models},
  author={Shen, Yi and Zhang, Jian and Huang, Jieyun and Shi, Shuming and Zhang, Wenjing and Yan, Jiangze and Wang, Ning and Wang, Kai and Liu, Zhaoxiang and Lian, Shiguo},
  booktitle={Proceedings of the 2025 Conference on Empirical Methods in Natural Language Processing: Industry Track},
  pages={2322--2331},
  year={2025}
}

@article{kim2023bigl,
  title={Speculative decoding with big little decoder},
  author={Kim, Sehoon and Mangalam, Karttikeya and Moon, Suhong and Malik, Jitendra and Mahoney, Michael W and Gholami, Amir and Keutzer, Kurt},
  journal={Advances in Neural Information Processing Systems},
  volume={36},
  pages={39236--39256},
  year={2023}
}

@article{chen2023frugalgpt,
  title={Frugalgpt: How to use large language models while reducing cost and improving performance},
  author={Chen, Lingjiao and Zaharia, Matei and Zou, James},
  journal={arXiv preprint arXiv:2305.05176},
  year={2023}
}

@article{ong2025routellm,
  title={Routellm: Learning to route llms with preference data},
  author={Ong, Isaac and Almahairi, Amjad and Wu, Vincent and Chiang, Wei-Lin and Wu, Tianhao and Gonzalez, Joseph E and Kadous, M Waleed and Stoica, Ion},
  journal={arXiv preprint arXiv:2406.18665},
  year={2024}
}

@inproceedings{muppidi2025predictive,
  title={Predictive Scheduling for Efficient Inference-Time Reasoning in Large Language Models},
  author={Muppidi, Aneesh and Brown, Katrina and Shahout, Rana},
  booktitle={ES-FoMo III: 3rd Workshop on Efficient Systems for Foundation Models},
  year={2025}
}

@article{korf1990rta,
  title={Real-time heuristic search},
  author={Korf, Richard E},
  journal={Artificial intelligence},
  volume={42},
  number={2-3},
  pages={189--211},
  year={1990},
  publisher={Elsevier}
}

@inproceedings{dean1988analysis,
  title={An Analysis of Time-Dependent Planning.},
  author={Dean, Thomas L and Boddy, Mark S and others},
  booktitle={AAAI},
  volume={88},
  pages={49--54},
  year={1988}
}

@article{zilberstein1996using,
  title={Using anytime algorithms in intelligent systems},
  author={Zilberstein, Shlomo},
  journal={AI magazine},
  volume={17},
  number={3},
  pages={73--73},
  year={1996}
}

@article{likhachev2003ara,
  title={ARA*: Anytime A* with provable bounds on sub-optimality},
  author={Likhachev, Maxim and Gordon, Geoffrey J and Thrun, Sebastian},
  journal={Advances in neural information processing systems},
  volume={16},
  year={2003}
}

@article{sutton1999between,
  title={Between MDPs and semi-MDPs: A framework for temporal abstraction in reinforcement learning},
  author={Sutton, Richard S and Precup, Doina and Singh, Satinder},
  journal={Artificial intelligence},
  volume={112},
  number={1-2},
  pages={181--211},
  year={1999},
  publisher={Elsevier}
}

@inproceedings{bacon2017optioncritic,
  title={The option-critic architecture},
  author={Bacon, Pierre-Luc and Harb, Jean and Precup, Doina},
  booktitle={Proceedings of the AAAI conference on artificial intelligence},
  volume={31},
  number={1},
  year={2017}
}

@article{baier2016time,
  title={Time management for Monte Carlo tree search},
  author={Baier, Hendrik and Winands, Mark HM},
  journal={IEEE transactions on computational intelligence and AI in games},
  volume={8},
  number={3},
  pages={301--314},
  year={2015},
  publisher={IEEE}
}

@inproceedings{huang2010timemanagement,
  title={Time management for Monte-Carlo tree search applied to the game of Go},
  author={Huang, Shih-Chieh and Coulom, Remi and Lin, Shun-Shii},
  booktitle={2010 International Conference on Technologies and Applications of Artificial Intelligence},
  pages={462--466},
  year={2010},
  organization={IEEE}
}

@article{baudivs2011pachi,
  title={Pachi: State of the art open source Go program},
  author={Baudi{\v{s}}, Petr and Gailly, Jean-loup},
  journal={Advances in computer games},
  pages={24--38},
  year={2011},
  publisher={Springer}
}

@article{russell1991architecture,
  title={An architecture for bounded rationality},
  author={Russell, Stuart J},
  journal={ACM SIGART Bulletin},
  volume={2},
  number={4},
  pages={146--150},
  year={1991},
  publisher={ACM New York, NY, USA}
}

@article{good1952rational,
  title={Rational decisions},
  author={Good, Irving John},
  journal={Journal of the Royal Statistical Society: Series B (Methodological)},
  volume={14},
  number={1},
  pages={107--114},
  year={1952},
  publisher={Wiley Online Library}
}

@article{simon1972theories,
  title={Theories of bounded rationality},
  author={Simon, Herbert A and others},
  journal={Decision and organization},
  volume={1},
  number={1},
  pages={161--176},
  year={1972},
  publisher={Amsterdam}
}

@phdthesis{anthony2021expert,
  title={Expert iteration},
  author={Anthony, Thomas William},
  year={2021},
  school={UCL (University College London)}
}

@article{horvitz2013reasoning,
  title={Reasoning, metareasoning, and mathematical truth: Studies of theorem proving under limited resources},
  author={Horvitz, Eric J and Klein, Adrian},
  journal={arXiv preprint arXiv:1302.4960},
  year={2013}
}

@inproceedings{horvitz1989reflection,
  title={Reflection and action under scarce resources: Theoretical principles and empirical study},
  author={Horvitz, Eric J and Cooper, Gregory F and Heckerman, David E},
  booktitle={IJCAI},
  volume={2},
  pages={1121--1127},
  year={1989}
}

@software{deepmind2020jax,
  title = {The {D}eep{M}ind {JAX} {E}cosystem},
  author = {DeepMind and Babuschkin, Igor and Baumli, Kate and Bell, Alison and Bhupatiraju, Surya and Bruce, Jake and Buchlovsky, Peter and Budden, David and Cai, Trevor and Clark, Aidan and Danihelka, Ivo and Dedieu, Antoine and Fantacci, Claudio and Godwin, Jonathan and Jones, Chris and Hemsley, Ross and Hennigan, Tom and Hessel, Matteo and Hou, Shaobo and Kapturowski, Steven and Keck, Thomas and Kemaev, Iurii and King, Michael and Kunesch, Markus and Martens, Lena and Merzic, Hamza and Mikulik, Vladimir and Norman, Tamara and Papamakarios, George and Quan, John and Ring, Roman and Ruiz, Francisco and Sanchez, Alvaro and Sartran, Laurent and Schneider, Rosalia and Sezener, Eren and Spencer, Stephen and Srinivasan, Srivatsan and Stanojevi\'{c}, Milo\v{s} and Stokowiec, Wojciech and Wang, Luyu and Zhou, Guangyao and Viola, Fabio},
  url = {http://github.com/deepmind},
  year = {2020},
}

@inproceedings{koyamada2023pgx,
  title={Pgx: Hardware-Accelerated Parallel Game Simulators for Reinforcement Learning},
  author={Koyamada, Sotetsu and Okano, Shinri and Nishimori, Soichiro and Murata, Yu and Habara, Keigo and Kita, Haruka and Ishii, Shin},
  booktitle={Advances in Neural Information Processing Systems},
  pages={45716--45743},
  volume={36},
  year={2023}
}

@article{schulman2017proximal,
  title={Proximal policy optimization algorithms},
  author={Schulman, John and Wolski, Filip and Dhariwal, Prafulla and Radford, Alec and Klimov, Oleg},
  journal={arXiv preprint arXiv:1707.06347},
  year={2017}
}

@article{schulman2015high,
  title={High-dimensional continuous control using generalized advantage estimation},
  author={Schulman, John and Moritz, Philipp and Levine, Sergey and Jordan, Michael and Abbeel, Pieter},
  journal={arXiv preprint arXiv:1506.02438},
  year={2015}
}

@misc{bonnet2024jumanji,
    title={Jumanji: a Diverse Suite of Scalable Reinforcement Learning Environments in JAX},
    author={Clément Bonnet and Daniel Luo and Donal Byrne and Shikha Surana and Sasha Abramowitz and Paul Duckworth and Vincent Coyette and Laurence I. Midgley and Elshadai Tegegn and Tristan Kalloniatis and Omayma Mahjoub and Matthew Macfarlane and Andries P. Smit and Nathan Grinsztajn and Raphael Boige and Cemlyn N. Waters and Mohamed A. Mimouni and Ulrich A. Mbou Sob and Ruan de Kock and Siddarth Singh and Daniel Furelos-Blanco and Victor Le and Arnu Pretorius and Alexandre Laterre},
    year={2024},
    eprint={2306.09884},
    url={https://arxiv.org/abs/2306.09884},
    archivePrefix={arXiv},
    primaryClass={cs.LG}
}

@book{puterman1994mdp,
  title={Markov Decision Processes: Discrete Stochastic Dynamic Programming},
  author={Puterman, Martin L.},
  year={1994},
  publisher={John Wiley \& Sons}
}

@article{forkel2025entropy,
  title={Entropy is all you need for Inter-Seed Cross-Play in Hanabi},
  author={Forkel, Johannes and Foerster, Jakob},
  journal={arXiv preprint arXiv:2511.22581},
  year={2025}
}

@article{muppidi2024fast,
  title={Fast trac: A parameter-free optimizer for lifelong reinforcement learning},
  author={Muppidi, Aneesh and Zhang, Zhiyu and Yang, Heng},
  journal={Advances in Neural Information Processing Systems},
  volume={37},
  pages={51169--51195},
  year={2024}
}

@article{boopathy2024permutation,
  title={Permutation Invariant Learning with High-Dimensional Particle Filters},
  author={Boopathy, Akhilan and Muppidi, Aneesh and Yang, Peggy and Iyer, Abhiram and Yue, William and Fiete, Ila},
  journal={arXiv preprint arXiv:2410.22695},
  year={2024}
}

@inproceedings{cope_learning_2023,
	title = {Learning to {Plan} with {Tree} {Search} via {Deep} {RL}},
	language = {en},
    booktitle = {{PRL Workshop Series} {\textendash} {Bridging the Gap Between AI Planning and Reinforcement Learning}},
	urldate = {2026-05-05},
	author = {Cope, Dylan and Svegliato, Justin and Russell, Stuart},
	year = {2023},
}

@article{otto2019opportunity,
  title={The opportunity cost of time modulates cognitive effort},
  author={Otto, A Ross and Daw, Nathaniel D},
  journal={Neuropsychologia},
  volume={123},
  pages={92--105},
  year={2019},
  publisher={Elsevier}
}

\appendix

\section{Reproducibility}
\label{app:reproducibility}

The full JAX implementation of every environment, base planner, and
gating policy in this paper, together with pretrained checkpoints for
all five environments and the two-GPU deployment harness, is released
at \url{https://aneeshers.github.io/realtime-rl/}.
Each result reported in Sections~\ref{sec:experiments}
and~\ref{sec:deployment} has a corresponding environment-variable-driven
launcher script (e.g.\ \texttt{eval\_pacman\_gating.sh},
\texttt{deploy.sh}) that reproduces the reported number from a single
command using the shipped checkpoints, and the release README lists
expected returns per command so any discrepancy is immediately visible.
Full training hyperparameters are in Appendix~\ref{app:implementation},
base-checkpoint selection is in Appendix~\ref{app:cross_eval}, and the
two-GPU deployment grid is in
Appendix~\ref{app:deployment_details}.

\section{AlphaZero and MCTS}
\label{app:az_primer}

This appendix explains the planning infrastructure used in this paper.
It targets readers familiar with deep RL but not with tree-search
methods.

The planner used throughout this paper is Monte Carlo Tree Search
(MCTS) guided by a neural network, the same combination introduced
in AlphaZero \citep{silver2018general}. The intuition is that the
network alone is fast but imperfect: it gives a quick estimate of which
actions look promising, but it never actually checks. MCTS turns a
fixed compute budget into action-quality: it spends that budget on
simulated rollouts, focusing on the actions the network considers most
promising while still exploring less-likely alternatives, and uses what
it learns to choose a better action than the network's raw guess. More
simulations produce better actions but take longer, which is the
tradeoff the gating policy in the main paper learns to manage.

\textbf{What does ``running a simulation'' mean in MCTS?}

A \emph{simulation} (or \emph{playout}) is one traversal of the search
tree from the root (the current state) down to a leaf, followed by a
backup of the resulting value estimate up the tree. Each step of the
traversal picks a child to descend into, balancing two pressures:
prefer children that have looked good so far (high empirical value),
but also try children the search has barely visited (high prior, low
visit count) in case those are even better. The PUCT rule formalizes
this:
\[
  a^* = \arg\max_a \Bigl[Q(s,a)
        + c \cdot \pi_\theta(a|s)
          \cdot \frac{\sqrt{N(s)}}{1+N(s,a)}\Bigr],
\]
where $Q(s,a)$ is the empirical mean value of subtree $a$,
$\pi_\theta(a|s)$ is the network's prior over actions,
$N(s)$ is the total visit count of node $s$,
and $N(s,a)$ is the visit count of edge $(s,a)$.
Repeated simulations concentrate visits on high-value, high-prior
subtrees while continuing to explore less-visited ones.
After $n$ simulations the recommended action is the
\emph{most-visited} child of the root.

\textbf{How is the final action selected?}

At evaluation time the agent picks the most-visited root child.
During training, the visit-count distribution
$\hat{\pi}(a) \propto N(s_0,a)^{1/\tau}$ (temperature $\tau > 0$)
is used as a \emph{search target}: the network is updated to predict
$\hat{\pi}$ at state $s_0$. Lower $\tau$ concentrates probability on
the most-visited action; higher $\tau$ spreads probability across
alternatives, useful for exploration early in training.
This setup (search produces a better-than-raw-policy distribution,
and the network learns to imitate it) is called \emph{expert
iteration} \citep{anthony2021expert}, and over many iterations it tightens the network's prior,
making future searches better-targeted.

\textbf{What is the difference between AlphaZero and MuZero?}

\textbf{AlphaZero}~\cite{silver2018general} requires a \emph{perfect
simulator}: given state $s$ and action $a$, the simulator returns the
exact next state $s'$. MCTS runs entirely inside this simulator; the
network supplies priors and value estimates but never models
transitions.

\textbf{MuZero}~\cite{schrittwieser2020muzero} replaces the perfect
simulator with a \emph{learned latent-space dynamics model}.
Transitions during search are predicted by a network rather than a
ground-truth engine, so MuZero can plan in environments without a
simulator (e.g.\ Atari frames).

All environments in this paper expose a JAX-native step function, so we
use the AlphaZero paradigm. This choice is also what enables our
committed-action protocol (Section~\ref{sec:committed}): the search
tree explicitly rolls forward the $k{-}1$ committed steps to reach the
future state where the planned action will land, which is only possible
when the simulator is perfect.

\textbf{AlphaZero was designed for two-player zero-sum games. How does
it transfer to single-agent settings?}

The core loop is \emph{expert iteration}~\cite{anthony2021expert}: at
each iteration the agent uses MCTS to produce an expert distribution
over actions (better than the network's raw policy because it
incorporates lookahead), and the network is trained to imitate that
expert. Over many iterations, the network's prior gets stronger, which
makes future searches better-targeted. The value head uses a
bootstrapped return instead of a zero-sum game outcome; otherwise the
algorithm is identical to two-player AlphaZero.
Algorithm~\ref{alg:expert_iter} formalizes this.

\begin{algorithm}[h]
\caption{Single-agent AlphaZero via Expert Iteration}
\label{alg:expert_iter}
\begin{algorithmic}[1]
\Require Network $f_\theta = (\pi_\theta, v_\theta)$, environment $\mathcal{E}$,
         simulations per step $n$, replay buffer $\mathcal{B}$
\Repeat
  \State $\mathcal{D} \leftarrow \{\}$ \Comment{episode data buffer}
  \State $s_0 \leftarrow \mathcal{E}.\text{reset}()$
  \For{$t = 0, 1, \ldots, T-1$}                     \Comment{\textbf{Act}}
    \State Run $n$ MCTS simulations from $s_t$ using $f_\theta$
    \State $\hat{\pi}_t(a) \propto N(s_t, a)^{1/\tau}$ \Comment{visit-count policy}
    \State $a_t \sim \hat{\pi}_t$;\quad $s_{t+1} \leftarrow \mathcal{E}.\text{step}(a_t)$
    \State $\mathcal{D} \leftarrow \mathcal{D} \cup \{(s_t,\, \hat{\pi}_t)\}$
  \EndFor
  \State Compute returns $z_t = \sum_{k \geq 0} \gamma^k r_{t+k}$ for each $t$ \Comment{\textbf{Store}}
  \State $\mathcal{B} \leftarrow \mathcal{B} \cup \{(s_t, \hat{\pi}_t, z_t)\}_{t=0}^{T-1}$
  \State Sample mini-batch from $\mathcal{B}$                     \Comment{\textbf{Learn}}
  \State $\mathcal{L}(\theta) =
    \displaystyle\frac{1}{|\mathcal{B}|}\sum_{(s,\hat{\pi},z)\in\mathcal{B}}
    \Bigl[\bigl(v_\theta(s) - z\bigr)^2
          - \hat{\pi}^\top \log \pi_\theta(s)\Bigr]$
  \State $\theta \leftarrow \theta - \alpha\,\nabla_\theta\,\mathcal{L}(\theta)$
\Until{convergence}
\end{algorithmic}
\end{algorithm}

\textbf{How does MCTS scale on modern accelerators?}

A common worry is that MCTS is inherently sequential, each
simulation updates the $Q$ and $N$ tables that the next simulation
reads. In practice, two layers of parallelism make this much less of a
bottleneck than it appears.

\textbf{Environment-level parallelism.}
We maintain a batch of $E$ independent environment instances (32 in our
experiments). Each instance has its own search tree; all $E$ leaf nodes
are evaluated in a \emph{single batched forward pass} of the neural
network, which is the GPU/TPU bottleneck.

\textbf{Intra-tree simulation loop via \texttt{jax.lax.scan}.}
Within each tree the $n$ simulations are unrolled as a \texttt{scan}.
Each iteration: (1) traverse all $E$ trees to their current leaf using
PUCT (parallelised over $E$), (2) batch-evaluate the $E$ leaves in one
network call, (3) back-propagate values. Because JAX traces the full
scan at compile time, the tree-update logic is fused into a single XLA
kernel with zero Python-level iteration overhead at runtime.

The result is that all $n$ simulations across all $E$ environments are
compiled into a single \texttt{jit}-ted function that runs entirely on
the accelerator, with one network forward pass per simulation step and
no host-device transfers during rollout.

\section{Variable-duration GAE}
\label{app:gae}

We re-introduce the meta-step subscript $k_t$ in this appendix because the
derivation reasons about a sequence of meta-decisions rather than a single
budget choice.

Standard GAE~\citep{schulman2015high} estimates the advantage at
timestep $t$ as a $\lambda$-weighted sum of one-step TD residuals,
\begin{equation}
  \hat{A}_t \;=\; \sum_{l=0}^{\infty} (\gamma \lambda)^l\, \delta_{t+l},
  \qquad
  \delta_t \;=\; r_t + \gamma\, V(s_{t+1}) - V(s_t),
\end{equation}
which assumes a unit time gap between consecutive states.
In our SMDP, consecutive meta-states $s_t$ and $s_{t + k_t}$ are
separated by a variable number of environment frames $k_t$, so the
per-step discount in the TD residual becomes $\gamma^{k_t}$:
\begin{equation}
  \delta_t \;=\; R_t + \gamma^{k_t}\, V(s_{t + k_t}) - V(s_t),
  \qquad
  R_t \;=\; \sum_{j=0}^{k_t - 1} \gamma^j\, r_{t+j}.
\end{equation}
The discount accumulated between meta-steps $t$ and $t+l$ is
$\gamma^{\sum_{j=0}^{l-1} k_{t+j}}$ rather than $\gamma^l$, giving the
meta-step advantage
\begin{equation}
  \hat{A}_t
  \;=\;
  \sum_{l=0}^{\infty} \lambda^l
  \left(\prod_{j=0}^{l-1} \gamma^{k_{t+j}}\right) \delta_{t+l},
\end{equation}
or, recursively,
\begin{equation}
  \hat{A}_t \;=\; \delta_t + \gamma^{k_t} \lambda\, \hat{A}_{t+1}.
  \label{eq:gae_recursive}
\end{equation}
The structure is identical to standard GAE; the only change is that
the per-step discount is $\gamma^{k_t}$ rather than $\gamma$.
We compute these advantages in a single backward pass over each
rollout using equation~\eqref{eq:gae_recursive}.

Without this correction, treating each meta-step as a unit time gap
would systematically understate the discount on long-budget
meta-steps, biasing the value function in favor of many short budgets
over fewer long ones.
The $\gamma^{k_t}$ correction makes the value function comparable
across budget choices.

\section{Environment Details}
\label{app:envs}

\subsection{Real-time Tetris}
\label{app:envs_tetris}

Our real-time Tetris environment is a modification of Jumanji Tetris in which each
environment step is one gravity tick rather than one full placement decision.
The board is $20 \times 10$, episodes last at most 2000 gravity ticks, and the
observation contains the locked board, the current falling tetromino, its
$(x,y)$ position, and the gravity-tick count.

Every call to \texttt{env.step(action)} first applies the chosen control and then
applies one gravity update. The action set has six elements:
\begin{itemize}[nosep]
  \item \textbf{Left / Right}: translate the falling piece one column in the respective
    direction; invalid lateral moves are ignored.
  \item \textbf{Rotate-CW / Rotate-CCW}: rotate the piece $90°$ clockwise or
    counter-clockwise; invalid rotations are ignored.
  \item \textbf{Hard-drop}: instantly translate the piece to the lowest valid row and
    lock it; a new piece spawns immediately.
  \item \textbf{Noop}: no horizontal or rotational change; the piece descends one row
    by gravity.
\end{itemize}
A piece also locks whenever gravity would move it into an occupied cell; completed
rows are then cleared and scored with the standard convex Tetris reward schedule.
The episode ends on board top-out, i.e.\ when a newly spawned piece cannot be placed
at the spawn position, or when the 2000-tick horizon is reached.

For the committed-action experiments in the main paper we use the \texttt{TetrisRTKT}
variant. The underlying environment dynamics are identical, but during the $k{-}1$
delay steps inside MCTS the tree rolls out policy-guided committed actions rather
than assuming a pure gravity-only noop sequence. This makes the search-time delay
model match the deployment-time reflex behavior more closely.

\subsection{Committed-action execution}
\label{app:committed_action}

During the $k{-}1$ delay steps while the MCTS planner is running, the agent executes
committed actions in the environment.
In all variants these are drawn from the argmax of the base model's raw policy
logits---a single inexpensive forward pass, with no additional MCTS compute.
The two variants differ only in what the \emph{MCTS tree simulates} during its internal
rollout of the delay window, not in what is executed in the environment.
We refer to these internal simulator variants as RT\_KStep (noop) and
RT\_KT (logit).

\textbf{RT\_KStep (noop)} simulates the delay steps inside the tree using a fixed
action (gravity/noop), which is fast but mismatches what the agent will actually do.

\textbf{RT\_KT (logit)} simulates the delay steps inside the tree using
$\mathrm{argmax}(\pi_{\mathrm{logits}})$, matching the committed actions the agent
executes at deployment and making the search model more accurate at no extra
environment cost.
All real-time Tetris results reported in the main paper use the RT\_KT variant.

Crucially, neither variant uses MCTS for the committed actions executed in the
environment, so there is no compute-parity violation: every option $k \in \Kset$
uses the same number of policy forward passes for its delay steps.

\subsection{Speed Hex}
Speed Hex is built on the pgx Hex library (11$\times$11 board).
Each player begins with $T = 300$ clock ticks; time spent planning is deducted after
each move.
MCTS search uses board-only dynamics (no clock deduction during simulation), but the
observation (and therefore the network's value estimates) includes the remaining
clock for both players.
The gating policy is a GRU network that maintains hidden state across moves within an
episode, allowing it to track game progression and clock pressure jointly.

\paragraph{Training notes.}
Early experiments fine-tuned the gating policy against fixed-budget opponents rather
than via self-play.
This regime exhibited plasticity loss, the network's ability to adapt to new opponents
degraded over successive fine-tuning rounds, consistent with findings in continual RL
settings~\citep{muppidi2024fast, boopathy2024permutation}.
We therefore switched to self-play, which eliminated plasticity loss entirely.
In self-play we found that increasing entropy regularization was the single most
impactful training intervention, consistent with recent results in other self-play
board-game domains~\citep{forkel2025entropy}.

\subsection{Speed Go}
Speed Go is built on the pgx Go library (9$\times$9 board).
As in Speed Hex, each player has an explicit clock, planning consumes clock only
after the move is played, and MCTS search itself uses board-only dynamics while the
network observes the remaining clock for both players.
We use the same recurrent gating architecture and evaluate the policy over the same
shared five sampled clock budgets used in the main text.
The same training observations apply: self-play with elevated entropy regularization
was decisive, and no plasticity loss was observed once fine-tuning against fixed
opponents was abandoned.

\section{Simulation Option Calibration for Clock Environments}
\label{app:sim_calibration}

For committed-action environments, the equal-budget constraint (32 sims/frame) naturally
produces meaningful option spacing.
For clock environments this constraint does not apply, so we select simulation options
by examining two curves simultaneously: planning quality (win rate or return) vs.\
simulation count, and inference latency vs.\ simulation count.

An option is \emph{useful} if it is distinguishable from its neighbors on both dimensions:
adding more simulations should noticeably improve play \emph{and} noticeably increase
latency.
Options that are close in both cost and quality give the gating policy nothing to learn;
options that differ in cost but not quality waste planning budget.

\begin{figure}[h]
  \centering
  \includegraphics[width=\linewidth]{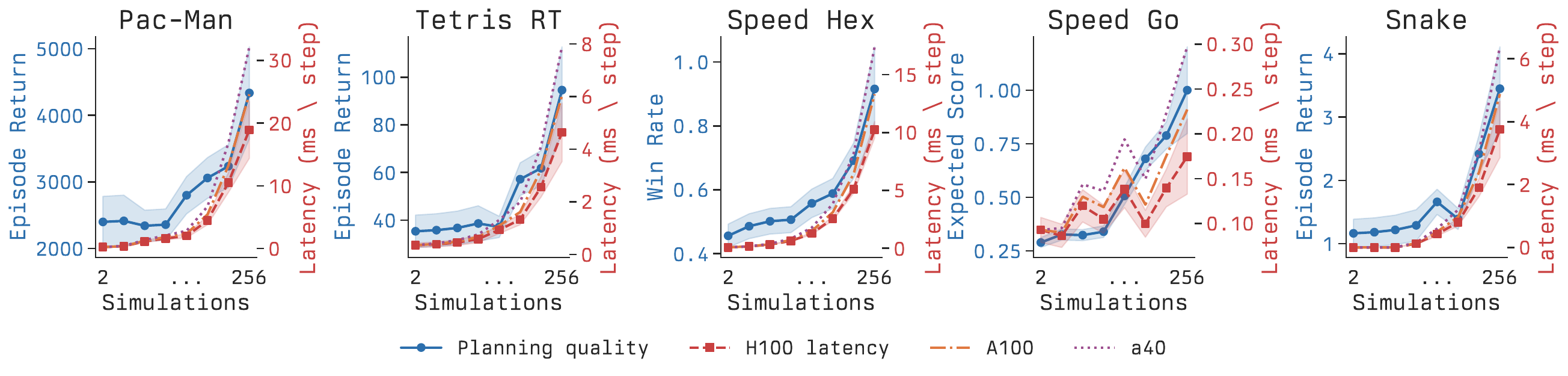}
  \caption{Full co-scaling results across Pac-Man, real-time Tetris, Speed
    Hex, Speed Go, and Snake. Blue denotes planning quality; dashed curves
    denote per-step latency on H100, A100, and A40; shading shows $\pm$SE.}
  \label{fig:scaling_appendix}
\end{figure}

For Speed Hex we found that options $\{2, 8, 32, 128\}$ satisfy this criterion across
the relevant range of clock settings, and for Speed Go we found that
$\{16, 32, 64, 96\}$ yield similarly distinct quality-latency tradeoffs.
Figure~\ref{fig:scaling_appendix} shows the full co-scaling results across all five
evaluation environments.

\begin{figure}[h]
  \centering
  \includegraphics[width=\linewidth]{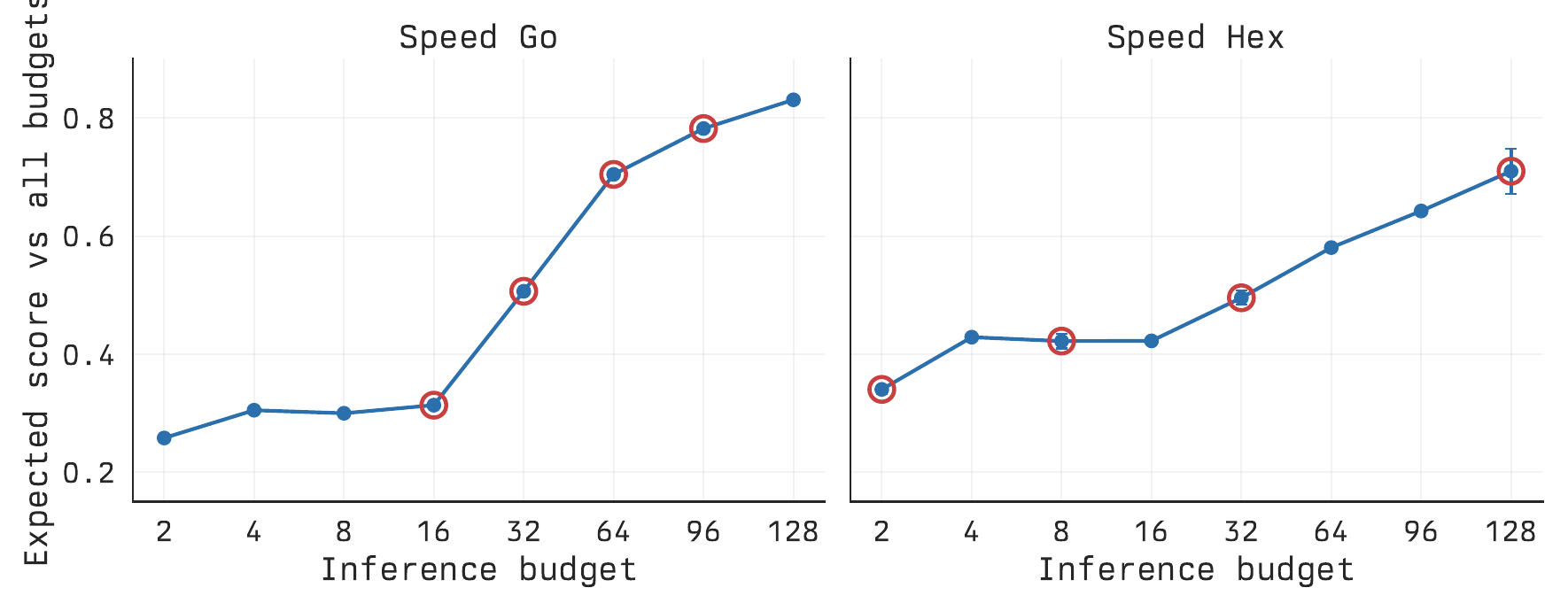}
  \caption{
    Simulation-option calibration for the two clock environments.
    Each point shows a budget's average expected score against all other
    candidate budgets.
    Red circles mark the options used in our clocked experiments:
    16/32/64/96 simulations for Speed Go and 2/8/32/128 for Speed Hex.}
  \label{fig:clock_budget_calibration}
\end{figure}

\paragraph{Go and Hex calibration.}
To calibrate the clock-environment options more directly, we examine how average
expected score changes as inference budget increases.
The resulting curves are shown in Figure~\ref{fig:clock_budget_calibration}.
For Speed Go, the selected set $\{16, 32, 64, 96\}$ spans the steep part of the
quality curve: average expected score versus all other budgets rises from $0.314$
at 16 simulations to $0.506$ at 32, $0.704$ at 64, and $0.782$ at 96.
The largest gains occur at $16{\rightarrow}32$ ($+0.193$) and
$32{\rightarrow}64$ ($+0.198$), while $64{\rightarrow}96$ still provides a clear
additional improvement ($+0.078$).
Pairwise tournament results among the selected Go options also remain well
separated: 64 simulations scores $0.713$ against 32 and $0.868$ against 16,
while 96 scores $0.595$ against 64 and $0.773$ against 32.

For Speed Hex, the selected set $\{2, 8, 32, 128\}$ follows the same design
principle and is now supported by the completed inference-budget tournament.
Average expected score versus all other budgets rises from $0.340$ at 2
simulations to $0.422$ at 8, $0.495$ at 32, and $0.710$ at 128, with successive
gains of $+0.082$, $+0.073$, and $+0.214$.
The pairwise Hex tournament shows the same separation: 8 simulations scores
$0.550$ against 2, 32 scores $0.564$ against 8 and $0.643$ against 2, and 128
scores $0.693$ against 32 and $0.727$ against 8.
Thus, in both clocked domains, the selected options provide the
gating policy with meaningfully separated quality-latency tradeoffs.

\section{Cross-Evaluation and Base Model Selection}
\label{app:cross_eval}

For the committed-action environments, we trained four AlphaZero checkpoints at
budgets $k \in \Kset$ and evaluated all 16 (train-$k$, eval-$k$) combinations.
Here, train-$k$ denotes the action-delay budget used during training, while
eval-$k$ denotes the action-delay budget used at test time.
For each train-$k$ checkpoint we additionally swept the MCTS simulation budget at
evaluation time and report the best raw episode return achieved for each eval-$k$.
The evaluation sweeps were: real-time Tetris with simulation counts
$\{32, 64, 96, 128\}$, Pac-Man with $\{8, 16, 32, 64, 96, 128\}$, and Snake
with $\{32, 64, 96, 128\}$.

\begin{table}[h]
  \centering
  \small
  \setlength{\tabcolsep}{5pt}
  \begin{tabular}{llcccc}
    \toprule
    Environment & Train-$k$ & Eval-$k{=}1$ & Eval-$k{=}2$ & Eval-$k{=}3$ & Eval-$k{=}4$ \\
    \midrule
    \multirow{4}{*}{real-time Tetris}
      & $k{=}1$ & 61.8 @ 128 & \textbf{85.0 @ 128} & \textbf{73.2 @ 128} & \textbf{69.0 @ 128} \\
      & $k{=}2$ & 30.0 @ 96  & 11.4 @ 128          & 7.4 @ 96            & 3.6 @ 96 \\
      & $k{=}3$ & \textbf{72.2 @ 128} & 22.0 @ 128 & 15.2 @ 128 & 11.8 @ 128 \\
      & $k{=}4$ & 69.6 @ 128 & 24.0 @ 96           & 23.2 @ 128          & 18.8 @ 128 \\
    \midrule
    \multirow{4}{*}{Pac-Man}
      & $k{=}1$ & \textbf{3235.4 @ 128} & 925.3 @ 128 & \textbf{2579.3 @ 128} & \textbf{1869.2 @ 96} \\
      & $k{=}2$ & 2458.5 @ 128 & 630.7 @ 128 & 866.7 @ 128 & 574.4 @ 96 \\
      & $k{=}3$ & 2749.7 @ 128 & 1122.3 @ 128 & 1110.6 @ 64 & 1051.2 @ 128 \\
      & $k{=}4$ & 2904.1 @ 128 & \textbf{1161.2 @ 128} & 1117.2 @ 128 & 1132.4 @ 96 \\
    \midrule
    \multirow{4}{*}{Snake}
      & $k{=}1$ & 0.77 @ 128 & 0.33 @ 128 & 0.50 @ 128 & 0.27 @ 128 \\
      & $k{=}2$ & 2.32 @ 128 & 0.56 @ 128 & 0.84 @ 128 & 0.31 @ 96 \\
      & $k{=}3$ & \textbf{2.42 @ 128} & \textbf{0.80 @ 64} & \textbf{1.13 @ 128} & \textbf{0.49 @ 96} \\
      & $k{=}4$ & 0.05 @ 32  & 0.05 @ 128 & 0.06 @ 128 & 0.07 @ 128 \\
    \bottomrule
  \end{tabular}
  \caption{
    Cross-evaluation of committed-action base models.
    Entries report the best raw episode return for each train-$k$/eval-$k$
    combination after optimizing over the evaluation simulation-budget sweep;
    the selected simulation count is shown after the `@' symbol. Bold marks the
    best train-$k$ for each eval-$k$ within an environment.}
\end{table}

The resulting pattern is environment-dependent but clear.
For real-time Tetris, $k{=}1$ is the strongest base model overall: although the
$k{=}3$ checkpoint is best at eval-$k{=}1$, the $k{=}1$ checkpoint wins at
eval-$k \in \{2,3,4\}$ and has the highest mean return when averaging the
best-per-eval-$k$ scores.
For Pac-Man, $k{=}1$ also wins overall, dominating eval-$k \in \{1,3,4\}$ and
achieving the highest average performance across eval budgets, while $k{=}4$
is only slightly better at eval-$k{=}2$.
Snake differs qualitatively: the $k{=}3$ checkpoint is the strongest across all
four eval budgets and therefore serves as the frozen base model in that domain.

These results match the intuition that low-delay training is often causally
cleaner, since less delay-induced mismatch enters the value targets.
In addition, committed actions during delay steps are drawn from the base
model's policy, so a stronger base policy also produces better intermediate
states before the final MCTS action lands.
Accordingly, we use the $k{=}1$ checkpoint as the frozen base for Pac-Man and
real-time Tetris, and the $k{=}3$ checkpoint for Snake.

\section{Strategy Profiles Across Environments}
\label{app:strategy_appendix}

\begin{figure}[t]
  \centering
  \includegraphics[width=\linewidth]{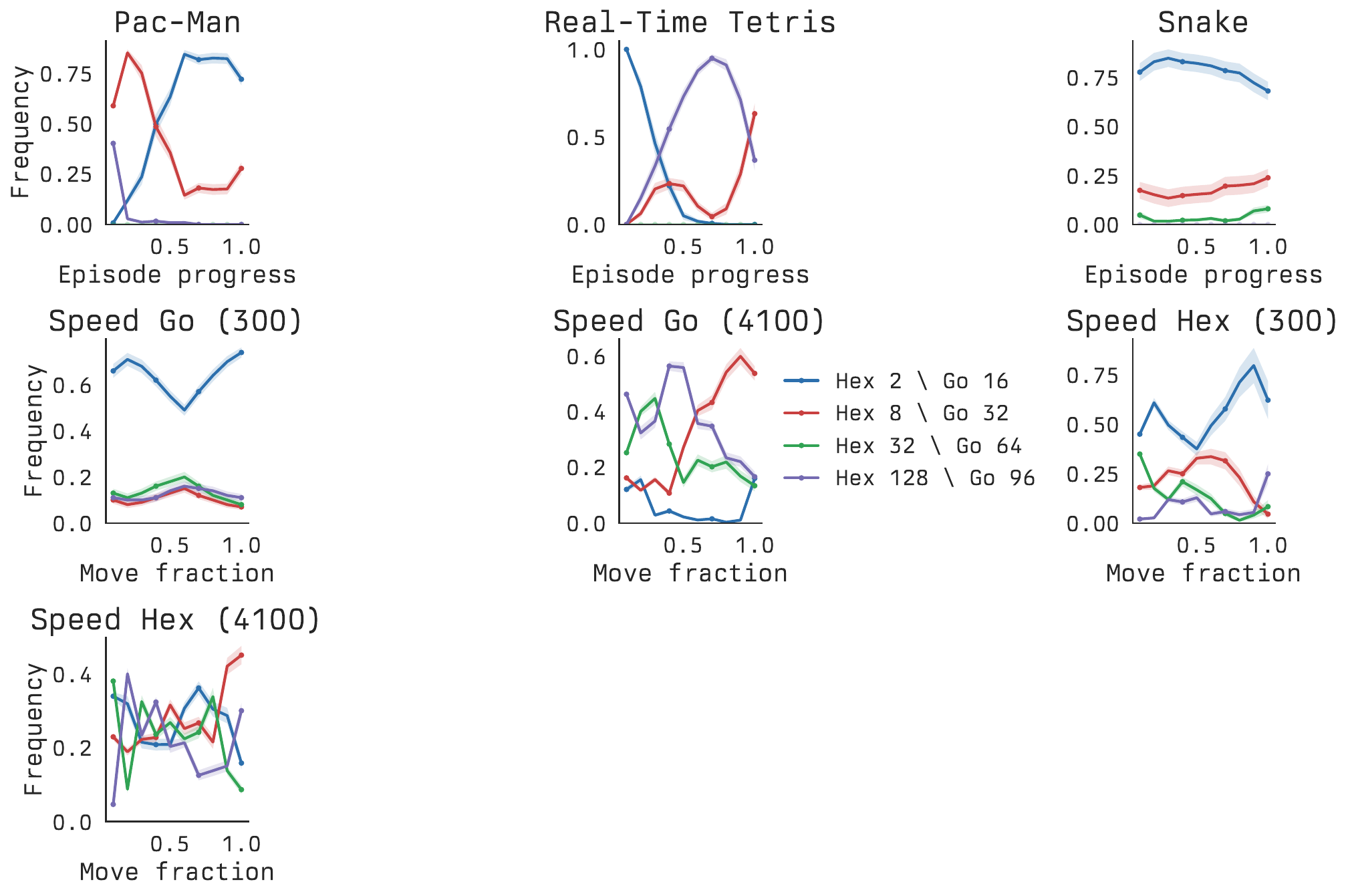}
  \caption{Different environments induce different allocation profiles. Pac-Man
    becomes more reactive later in the episode, real-time Tetris shifts toward deeper
    planning as the board densifies, Snake remains mostly reactive with occasional
    deeper planning in constrained states, and both clock games become less
    reactive when more time is available. This figure complements
    Figure~\ref{fig:strategy} by showing the full cross-environment strategy
    profiles rather than only the clock-game comparison.}
  \label{fig:strategy_appendix}
\end{figure}

This section expands the main-text analysis by showing the full learned budget-allocation profiles for each environment (Figure \ref{fig:strategy_appendix}).

\section{Strict-Timeout Speed Hex Control}
\label{app:hex_timeout_control}

The main Speed Hex setup allows the game to continue after the acting side exhausts
its remaining clock budget, so the experiment measures \emph{resource allocation}
rather than a pure timeout-avoidance game. To verify that this distinction matters,
we ran a strict-timeout control in which overspending the selected search budget
causes an immediate loss and there is no fallback behavior.

\begin{figure}[t]
  \centering
  \includegraphics[width=\linewidth]{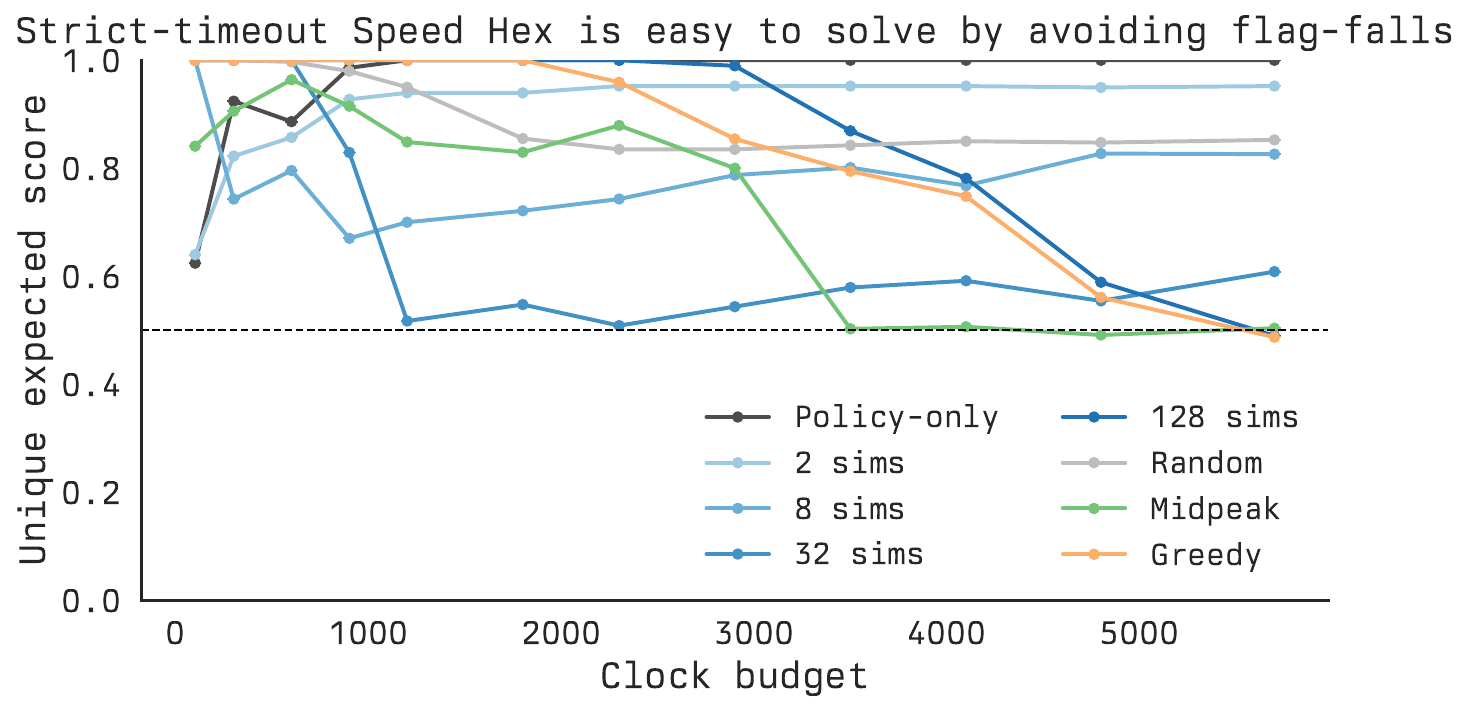}
  \caption{Strict-timeout Speed Hex is substantially easier than the main Speed Hex
    benchmark. Using the unique-game expected-score metric, the learned gate remains
    above parity against every opponent on average, with means of 0.952 against the
    policy-only opponent, 0.903 against fixed 2-simulation play, 0.904 against random
    allocation, and 0.893 even against the fixed 128-simulation opponent. At large
    budgets, the more aggressive opponents become the weakest because they overspend
    clock and lose on time.}
  \label{fig:hex_timeout_appendix}
\end{figure}

Figure~\ref{fig:hex_timeout_appendix} shows the resulting per-budget head-to-head
curves under the unique-game metric used in the main paper, evaluated on a wider
clock sweep than the main benchmark. The learned gate stays
above parity against every opponent \emph{on average}, and against five of the eight
opponents it is above parity at \emph{every} evaluated budget. Even the hardest
fixed-budget baseline under this metric, always-32, averages only 0.690 against the
gate, while policy-only averages 0.952 and random allocation averages 0.904. The
only sub-parity points occur for the more clock-aggressive baselines at the largest
budgets: fixed 128-simulation play dips to 0.490 at $T{=}5700$, midpeak dips to
0.491 at $T{=}4100$, and greedy dips to 0.487 at $T{=}5700$. This pattern is the
opposite of what we observe in the main benchmark. Once timeout is an immediate
terminal event, stronger search is no longer reliably beneficial; it often just
increases the probability of burning too much clock. In other words, the game becomes
easier because the gating policy can win largely by learning a conservative
time-management rule rather than by solving the underlying Hex positions more
effectively.

\section{Implementation Details}
\label{app:implementation}

\subsection{Gating network architecture}

\textbf{Spatial branch.}
The game grid is processed by a $1{\times}1$ convolution lifting to 64 channels,
followed by three residual blocks with Layer Normalization and global average pooling
to a 64-dimensional spatial representation.
Layer Normalization is used because the gating policy operates on a single environment
instance per rollout step, making batch statistics undefined.

\textbf{Time / clock embedding.}
For committed-action environments, the normalized step count is encoded as a 2-vector
and projected through a small MLP before being added to the spatial features.
For clock environments, the remaining clock fraction is similarly embedded and added.

\textbf{Fusion and heads.}
The spatial representation (64-dim), planner trunk features (128-dim), and planner
value (1-dim) are concatenated and passed through dual MLP heads: a policy head
producing logits over $k$, and a value head producing a scalar baseline.

\textbf{Speed Hex gating network.}
Uses a GRU backbone rather than a feedforward network, maintaining a recurrent hidden
state across moves within each game to track temporal context.

\subsection{Training hyperparameters}

\begin{table}[h]
  \centering
  \caption{PPO hyperparameters for committed-action environments.}
  \label{tab:hparams}
  \begin{tabular}{lccc}
    \toprule
    Hyperparameter & Pac-Man & real-time Tetris & Snake \\
    \midrule
    Num.\ parallel envs     & 32   & 32   & 32   \\
    Meta-steps per rollout  & 384  & 384  & 384  \\
    PPO epochs per rollout  & 4    & 4    & 4    \\
    Num.\ mini-batches      & 16   & 16   & 16   \\
    Meta-reward mode        & raw  & raw  & raw  \\
    Discount $\gamma$       & 0.99 & 0.99 & 0.997 \\
    GAE $\lambda$           & 0.95 & 0.95 & 0.95 \\
    PPO clip $\epsilon$     & 0.2  & 0.2  & 0.2  \\
    Entropy coefficient     & 0.01 & 0.01 & 0.05 \\
    Learning rate           & 3e-4 & 3e-4 & 3e-4 \\
    Sim options             & 32, 64, 96, 128 & 32, 64, 96, 128 & 32, 64, 96, 128 \\
    \bottomrule
  \end{tabular}
\end{table}

\paragraph{Clock-environment training details.}
The clock-game training runs use a separate self-play stack built on
\texttt{pgx}. For Speed Go, the base AlphaZero planner is trained by
pure self-play on 1024 parallel games with 32 MCTS simulations per
move, a 256-move horizon, training batch size 4096, Adam with
learning rate $10^{-3}$, and 800 training iterations; evaluation
against the \texttt{pgx} baseline occurs every 5 iterations and
checkpoints are saved every 200 iterations. The recurrent Speed Go
gate then loads the frozen 16-simulation base checkpoint matched to
the same seed and is trained in pure self-play for 500 PPO updates on
256 parallel environments. Each PPO update collects 32768 rollout
steps (128 per environment), chunks them into GRU sequences of length
64, and uses 4 PPO epochs with batch size 512, $\gamma{=}0.99$,
$\lambda{=}0.95$, PPO clip $\epsilon{=}0.2$, learning rate $3\times
10^{-4}$, value-loss coefficient 0.5, entropy coefficient 0.01, and
global gradient clipping at 1.0. The gate chooses among simulation
options $\{16,32,64,96\}$, domain-randomizes episode clocks over
$\{100,300,600,900,1200,1800,2300,2900,3500,4100,4800,5700\}$, and
saves checkpoints every 25 updates. Although the codebase contains
hooks for richer opponent schedules, the experiments reported here use
pure self-play throughout.

\subsection{JAX implementation and batching strategy}

The main computational challenge is that our system nests a meta-policy on top of
an AlphaZero-style planner, so a naive implementation would repeatedly invoke MCTS
inside an outer RL loop with substantial Python overhead. Our implementation avoids
this by pushing nearly all control flow into JAX primitives.

\textbf{Base planner training.}
The frozen planners are trained with our own Gumbel-AlphaZero stack built on top of
the Jumanji environments, using
\texttt{mctx.gumbel\_muzero\_policy} \citep{deepmind2020jax} for search. Within each planner step, the MCTS
simulations are executed inside a compiled \texttt{lax.scan}; leaf expansion and
backup are batched across parallel environments with \texttt{vmap}; and the full
self-play/training pipeline is wrapped in \texttt{jit}. In the multi-device setting,
the leading batch dimension is sharded with \texttt{pmap}, so each accelerator runs
the same compiled self-play/update program on its local slice of environments.

\textbf{Gating-policy training.}
For the gating layer, we again compile whole rollouts rather than individual steps.
Each PPO rollout is one \texttt{jit}-compiled \texttt{lax.scan} over meta-decisions,
and advantage computation is another reverse \texttt{lax.scan}. The recurrent Speed
Hex gate uses the same pattern: the GRU unroll is expressed as a scan, while PPO
operates on sequence chunks rather than on Python loops over timesteps.

\textbf{Evaluating all budget options in parallel.}
The key batching trick is that, during gating training and evaluation, we compute
the transition for every available budget option at each meta-step and only then
select the branch corresponding to the gate's chosen action. Concretely, for
$k \in \{1,2,3,4\}$ we run all four \texttt{meta\_step\_one\_k} branches, stack the
resulting rewards, discounts, next states, and next observations, and use a batched
selection operator to pick the chosen branch per environment. This spends extra
compute relative to evaluating only the selected option, but it preserves static
shapes, keeps the whole computation batched, and avoids recompilation or Python-side
control flow. In practice this systems tradeoff is what makes the meta-MCTS stack
fast enough to train routinely rather than as a one-off systems effort.

\textbf{Practical effect.}
These engineering choices --- \texttt{mctx} \citep{deepmind2020jax} for search, \texttt{vmap} for environment
parallelism, \texttt{jit} for whole-program compilation, \texttt{pmap} for device
parallelism, and \texttt{lax.scan} for both MCTS inner loops and PPO rollout loops ---
reduce wall-clock training time dramatically. In our internal runs, the end-to-end
training workflow dropped from roughly two weeks in an earlier less-batched pipeline
to roughly six hours in the optimized JAX implementation. The full implementation is
available at \url{https://aneeshers.github.io/realtime-rl/}.

\section{Detailed Two-GPU Deployment Breakdown}
\label{app:deployment_details}

\begin{figure}[h]
  \centering
  \includegraphics[width=\linewidth]{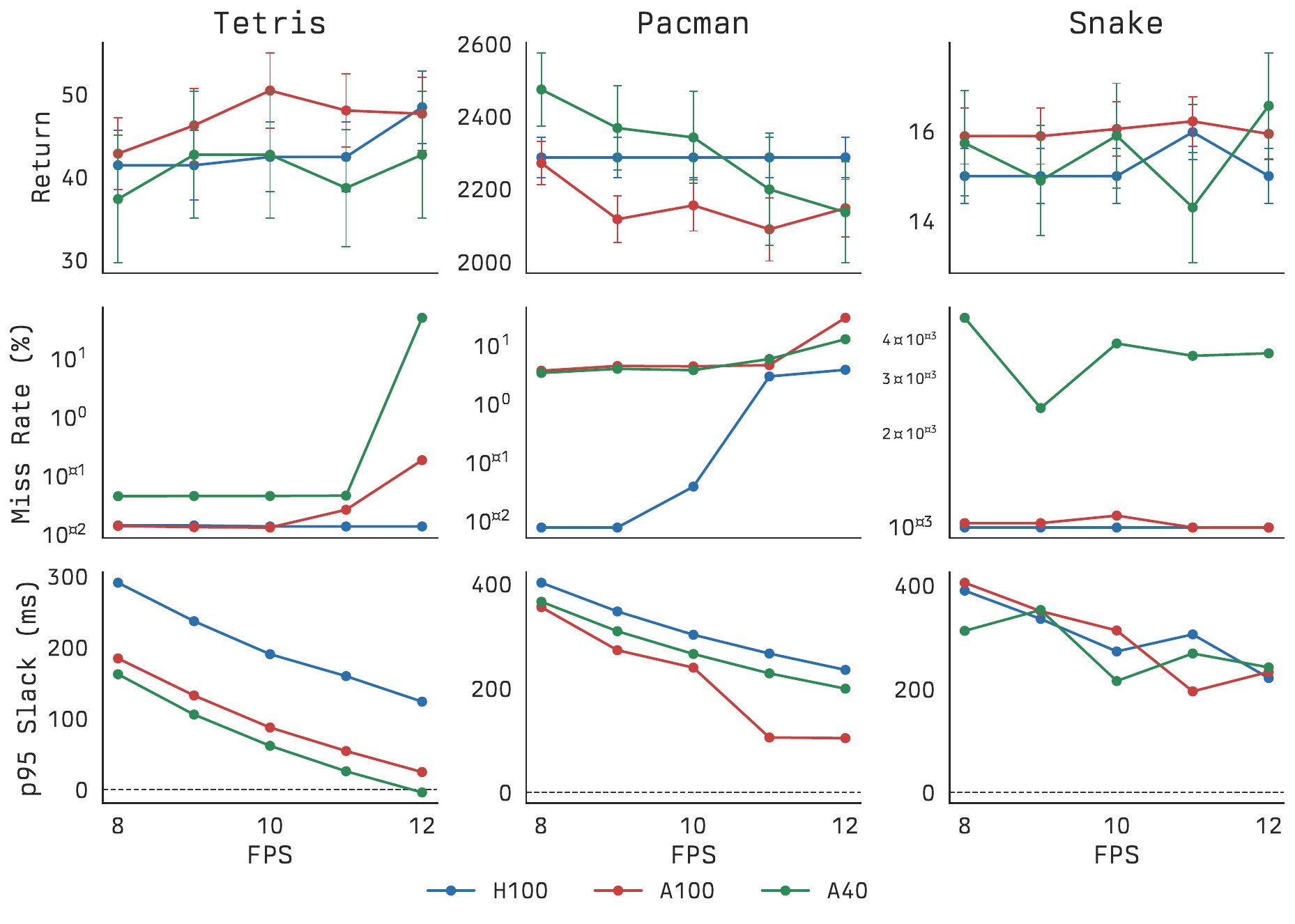}
  \caption{Detailed two-GPU deployment breakdown across real-time Tetris, Pac-Man, and Snake.
    Top: return vs.\ FPS. Middle: deadline miss rate. Bottom: p95 slack to the
    $k{=}4$ deadline. Snake remains robust, real-time Tetris fails only near the tightest
    A40 regime, and Pac-Man is the most latency-sensitive.}
  \label{fig:deployment_appendix}
\end{figure}

We ran the full asynchronous deployment grid for all 45 settings:
3 environments (real-time Tetris, Pac-Man, Snake) $\times$ 3 GPUs (H100, A100, A40)
$\times$ 5 frame rates (8--12\,FPS).
Figure~\ref{fig:deployment_appendix} gives the complete breakdown by environment,
GPU, and FPS for return, deadline miss rate, and p95 slack to deadline.

Several patterns are clear.
First, H100 is consistently reliable across the entire tested range.
In real-time Tetris its miss rate stays at $0.014\%$ for all FPS values, with p95 slack
remaining positive from $+291.0$\,ms at 8\,FPS to $+123.6$\,ms at 12\,FPS.
Snake is even easier: H100 and A100 stay at $0.001\%$ miss rate throughout, and
A40 stays at or below $0.005\%$.

Second, the deployment boundary appears where slack collapses.
Real-time Tetris on A40 is the clearest example: p95 slack decreases monotonically from
$+162.2$\,ms (8\,FPS) to $-4.1$\,ms (12\,FPS), and the miss rate jumps from
$0.046\%$ to $50.669\%$.
By contrast, real-time Tetris on A100 remains usable even at 12\,FPS, with p95 slack still
positive at $+24.5$\,ms and miss rate only $0.188\%$.

Third, Pac-Man is the most timing-sensitive environment.
Even with positive p95 slack throughout, A100 and A40 exhibit non-trivial miss
rates across all FPS values.
At 12\,FPS, Pac-Man reaches $30.369\%$ misses on A100 and $13.073\%$ on A40,
whereas H100 stays much lower at $3.927\%$.
This indicates that for Pac-Man, the far latency tail beyond the p95 threshold
matters operationally even when the central mass remains in-budget.

Finally, return remains comparatively stable over a broad deployment range despite
these timing differences.
Real-time Tetris returns remain in the $41$--$50$ range on H100/A100 and in the high-30s to
low-40s on A40 except at the most constrained settings.
Pac-Man degrades more gradually with hardware and FPS than its miss-rate curve might
suggest, while Snake is nearly flat across all tested settings.
Together these results support the main-text conclusion: the committed-action
training protocol transfers robustly to asynchronous hardware deployment, and the
observed failures emerge in predictable deadline-limited regimes rather than as
qualitatively new behaviors.

\section{Additional Results and Ablations}
\label{app:ablations}

We also ran input-ablation evaluations for the committed-action environments, in
which the gating policy received zeroed observation features, zeroed time
features, zeroed planner trunk features, or a zeroed planner value input while
the environment dynamics and MCTS planner remained unchanged.
Across environments, ablating planner trunk features was consistently
disruptive, while zeroing the scalar value input had only minor effect; in
real-time Tetris, zeroing either the board observation or planner trunk substantially
reduced return, while in Pac-Man and Snake the learned behavior was more robust
to removing any single input channel.

\begin{table}[h]
  \centering
  \small
  \setlength{\tabcolsep}{6pt}
  \begin{tabular}{lccccc}
    \toprule
    Environment & Baseline & Zero obs & Zero time & Zero trunk & Zero value \\
    \midrule
    Pac-Man   & 2245.6 & 2379.2 & 1571.2 & 1812.6 & 2189.8 \\
    real-time Tetris &   44.0 &   34.4 &   37.6 &   33.2 &   44.0 \\
    Snake     &   15.18 &  14.92 &  15.94 &  15.66 &  16.14 \\
    \bottomrule
  \end{tabular}
  \caption{
    Input-ablation results for the committed-action gating policies.
    Entries report mean episode return under each ablation condition while the
    environment and MCTS planner remain unchanged.}
\end{table}


\section{Real-Time RL, SMDPs, and Committed-Action Training}
\label{app:rtrl_smdp}

This appendix expands Section~\ref{sec:smdp} by focusing on the distinction
between the discrete-time SMDP used in training and the continuous-time SMDP
approximated in deployment.

\subsection*{The RTRL framework and its single-step restriction}

Section~\ref{sec:smdp} already gives the RTMDP equation and explains why
Ramstedt et al.'s fixed one-step delay formalism is not the natural
representation for our variable-budget committed-action setting.
The additional point needed here is that, once delay is represented as
budgeted options, training and deployment instantiate two closely related
SMDPs with different time models.

\subsection*{Training is a discrete SMDP; deployment is a genuine SMDP}

\paragraph{Simulation.}
In training the environment is synchronous: MCTS runs in zero simulated time.
As described in Section~\ref{sec:smdp}, each meta-step proceeds by choosing
$k$, executing $k{-}1$ committed argmax steps, and then applying the MCTS
result. From the meta-policy's perspective this is a
\emph{Semi-Markov Decision Process}~\citep{puterman1994mdp} with
\emph{deterministic} holding time $\tau = k$, so successive meta-decisions are
spaced $k$ environment steps apart and the appropriate discount is $\gamma^k$.
This is the discrete-time SMDP solved during training.

\paragraph{Deployment.}
In real-time deployment the holding time between meta-decisions equals
$k \times T_{\mathrm{frame}}$: MCTS runs concurrently on GPU~1 while the
environment executes the $k$ reflex frames on GPU~0, so the two durations
\emph{overlap} rather than sum.
Given $k$, the elapsed wall-clock time is $k \times T_{\mathrm{frame}}$
regardless of MCTS latency, provided search completes within the deadline.
The holding time is stochastic only because $k$ is chosen stochastically
by $\pigate$; this is a true SMDP in the continuous-time
sense~\citep{puterman1994mdp}.
Our measurements confirm that MCTS latency stays well within the frame budget\footnote{Note the distinction between \emph{frame budget} and \emph{planning budget} $k$. The frame budget is the time between frames at the chosen FPS (e.g., 111,ms at 9,FPS)}
(H100: mean $122$\,ms, p95 $205$\,ms; A40: mean $197$\,ms, p95 $338$\,ms
vs.\ a $k{=}4$ frame budget of $444$\,ms at 9\,FPS), so the deterministic
simulation discount $\gamma^k$ matches the deployment discount closely,
which is why sim-to-real transfer succeeds without retraining.

\subsection*{How committed-action training avoids state augmentation}

The central reason the $k$-step RTMDP state augmentation is unnecessary in
our framework is that \emph{the committed action at each intermediate step is
computed from the current state, not from a state $k$ steps ago}.
Specifically, the committed action at frame $t$ is
$a_t = \arg\max_a \pi_\theta(a \mid s_t)$, evaluated fresh in $\sim$2\,ms---well
within the $111$\,ms frame budget.
The MCTS search is initiated at $s_0$ (start of meta-step), but the tree
explicitly rolls forward the same $k{-}1$ committed steps that will be executed
in the environment and selects its terminal action for the resulting future
landing state. This delay is therefore explicitly modeled in training: the
$k{-}1$ committed steps actually execute in the environment before the MCTS
action lands, so the policy learns to handle the $k$-step gap without any state
augmentation.

In the RTRL vocabulary, the committed-action component operates in the
\emph{turn-based} regime (computation time $\ll$ frame time, so the environment
effectively pauses during action selection), while the MCTS component operates
in a $k$-step delay regime whose effects are absorbed by the training curriculum
rather than by augmenting the state space.
The key property that makes this possible---and that the standard RTMDP does not
exploit---is that the intermediate policy $\pi_{\mathrm{reflex}}$ is a
\emph{deterministic function of the current state}, so its actions do not need
to be carried in the state to recover the Markov property.


\end{document}